\documentclass{article}
\usepackage{microtype}
\usepackage{graphicx}
\usepackage{subfigure}
\usepackage{booktabs}
\usepackage{hyperref}

\usepackage[accepted]{icml2023}

\usepackage{amsmath}
\usepackage{amssymb}
\usepackage{mathtools}
\usepackage{amsthm}
\usepackage[capitalize,noabbrev]{cleveref}

\theoremstyle{plain}

\theoremstyle{definition}

\theoremstyle{remark}

\usepackage[textsize=tiny]{todonotes}

\usepackage{multirow}
\usepackage{float}
\usepackage{stfloats}

\renewcommand{\algorithmicrequire}{\textbf{Input:}}
\renewcommand{\algorithmicensure}{\textbf{Output:}}

\icmltitlerunning{Quantifying the Knowledge in GNNs for Reliable Distillation into MLPs}

\begin{document}

\twocolumn[
\icmltitle{Quantifying the Knowledge in GNNs for Reliable Distillation into MLPs}

\icmlsetsymbol{equal}{*}

\begin{icmlauthorlist}
\icmlauthor{Lirong Wu}{xxx}
\icmlauthor{Haitao Lin}{xxx}
\icmlauthor{Yufei Huang}{xxx}
\icmlauthor{Stan Z. Li}{xxx}
\end{icmlauthorlist}

\icmlaffiliation{xxx}{AI Lab, Research Center for Industries of the Future, Westlake University, Hangzhou, China}

\icmlcorrespondingauthor{Stan Z. Li}{stan.zq.li@westlake.edu.cn}
\icmlcorrespondingauthor{Lirong Wu}{wulirong@westlake.edu.cn}

\icmlkeywords{Machine Learning, ICML}

\vskip 0.3in
]

\printAffiliationsAndNotice{}

\begin{abstract}
To bridge the gaps between topology-aware Graph Neural Networks (GNNs) and inference-efficient Multi-Layer Perceptron (MLPs), GLNN \cite{zhang2021graph} proposes to distill knowledge from a well-trained teacher GNN into a student MLP. Despite their great progress, comparatively little work has been done to explore \emph{the reliability of different knowledge points (nodes) in GNNs, especially their roles played during distillation.} In this paper, we first quantify the knowledge reliability in GNN by measuring the invariance of their information entropy to noise perturbations, from which we observe that different knowledge points \emph{(1)} show different distillation speeds (\emph{temporally}); \emph{(2)} are differentially distributed in the graph (\emph{spatially}). To achieve reliable distillation, we propose an effective approach, namely \emph{Knowledge-inspired Reliable Distillation} (\texttt{KRD}), that models the probability of each node being an informative and reliable knowledge point, based on which we sample a set of additional reliable knowledge points as supervision for training student MLPs. Extensive experiments show that \texttt{KRD} improves over the vanilla MLPs by 12.62\% and outperforms its corresponding teacher GNNs by 2.16\% averaged over 7 datasets and 3 GNN architectures. Codes are publicly available at: \url{https://github.com/LirongWu/RKD}.
\end{abstract}

\vspace{-2em}
\section{Introduction}
Recent years have witnessed the great success of Graph Neural Networks (GNNs) \cite{hamilton2017inductive,wu2023beyond,velivckovic2017graph,liu2020towards,wu2020comprehensive,zhou2020graph,wu2021self,wu2021graphmixup} in handling graph-related tasks. Despite their great \emph{academic success}, Multi-Layer Perceptrons (MLPs) remain the primary workhorse for practical \emph{industrial applications}. One reason for such academic-industrial gap is the neighborhood-fetching latency incurred by data dependency in GNNs \cite{jia2020redundancy,zhang2021graph}, which makes it hard to deploy for latency-sensitive applications. Conversely, Multi-Layer Perceptrons (MLPs) involve no data dependence between data pairs and infer much faster than GNNs, but their performance is less competitive. Motivated by these complementary strengths and weaknesses, one solution to reduce their gaps is to perform GNN-to-MLP knowledge distillation \cite{yang2021extract,zhang2021graph,gou2021knowledge}, which extracts the knowledge from a well-trained teacher GNN and then distills the knowledge into a student MLP. 

\begin{figure}[!tbp]
	\begin{center}
		\includegraphics[width=0.9\linewidth]{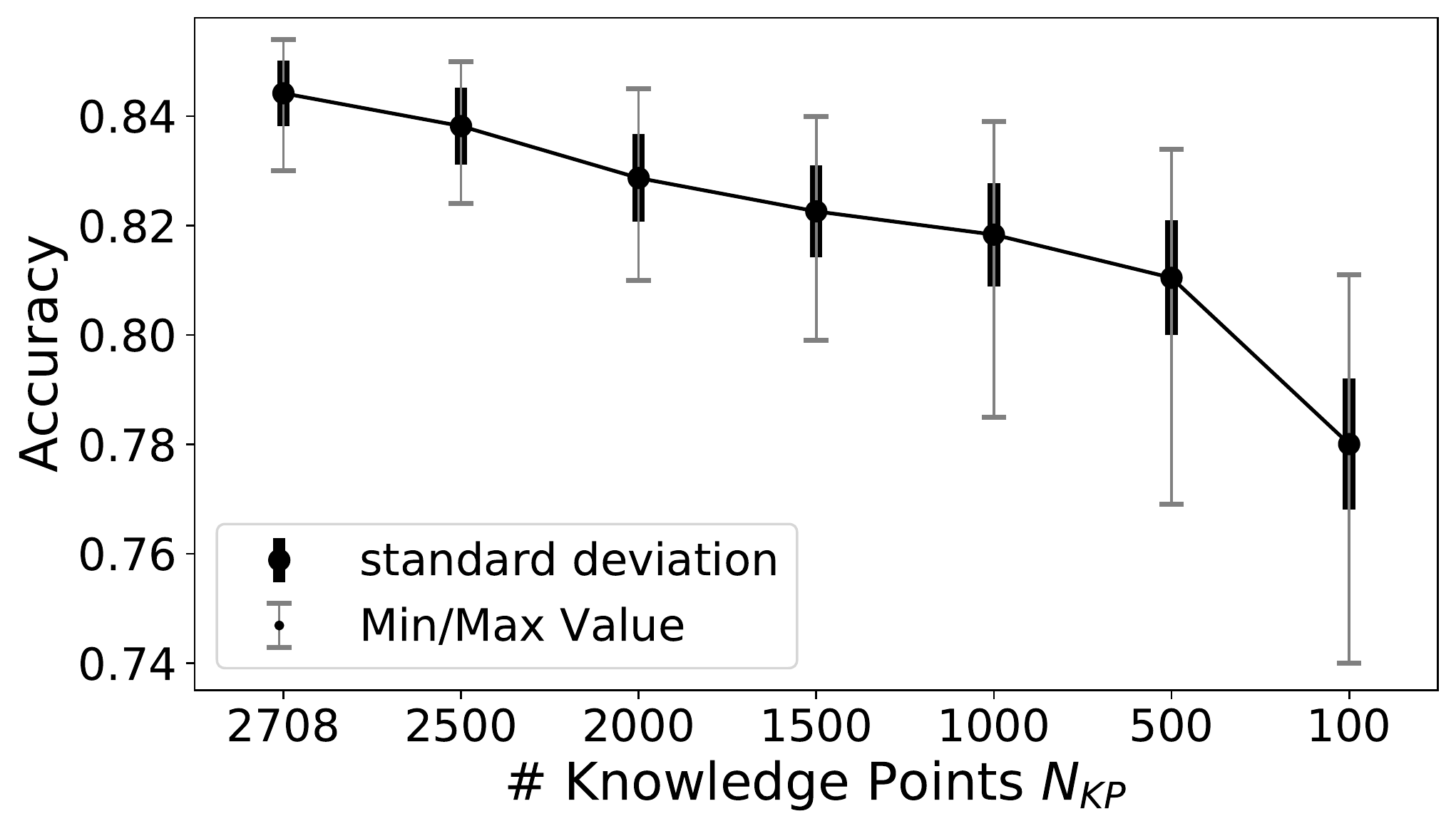}
	\end{center}
	\vspace{-1em}
	\caption{Mean, standard deviation, and minimum/maximum classification accuracy of student MLPs trained with different combinations of (randomly sampled) GNN knowledge points on \texttt{Cora}.}
	\vspace{-1em}
	\label{fig:1}
\end{figure}

Despite the great progress, most previous works have simply treated all knowledge points (nodes) in GNNs as equally important, and few efforts are made to explore \emph{the reliability of different knowledge points in GNNs and the diversity of the roles they play in the distillation process.} From the motivational experiment in Fig.~\ref{fig:1}, we can make two important observations about knowledge points: \textbf{\emph{(1) More is better:}} the performance of distilled MLPs can be improved as the number of knowledge points $N_{KP}$ increases; and \textbf{\emph{(2) Reliable is better:}} the performance variances (e.g., standard deviation and best/worst performance gap) of different knowledge combinations are enlarged as $N_{KP}$ decreases. The above two observations suggest that different knowledge points may play different roles in the distillation process and that distilled MLPs can consistently benefit from \emph{more reliable} knowledge points, while those uninformative and unreliable knowledge points may contribute little to the distillation.

\textbf{Present Work.} In this paper, we identify a potential \emph{under-confidence} problem for GNN-to-MLP distillation, i.e., the distilled MLPs may not be able to make predictions as confidently as teacher GNNs. Furthermore, we conduct extensive theoretical and experimental analysis on this problem and find that it is mainly caused by the lack of reliable supervision from teacher GNNs. To provide more supervision for reliable distillation into student MLPs, we propose to quantify the knowledge in GNNs by measuring the invariance of their information entropy to noise perturbations, from which we find that different knowledge points \emph{(1)} show different distillation speeds (\emph{temporally}); \emph{(2)} are differentially distributed in the graph (\emph{spatially}). Finally, we propose an effective approach, namely \emph{Knowledge-inspired Reliable Distillation} (\texttt{KRD}), for filtering out unreliable knowledge points and making full use of those with informative knowledge. The proposed \texttt{KRD} framework models the probability of each node being an information-reliable knowledge point, based on which we sample a set of additional reliable knowledge points as supervision for training student MLPs.

Our main contributions can be summarized as follows:

\vspace{-1em}
\begin{itemize}
    \item We are the first to identify a potential \emph{under-confidence} problem for GNN-to-MLP distillation, and more importantly, we described in detail what it represents, how it arises, what impact it has, and how to deal with it. 
    \item We propose a perturbation invariance-based metric to quantify the reliability of knowledge in GNNs and analyze the roles played by different knowledge nodes \emph{temporally} and \emph{spatially} in the distillation process. 
    \item We propose a \emph{Knowledge-inspired Reliable Distillation} (\texttt{KRD}) framework based on the quantified GNN knowledge to make full use of those reliable knowledge points as additional supervision for training MLPs.
\end{itemize}

\vspace{-1.5em}
\section{Related Work}
\vspace{-0.3em}
\noindent \textbf{GNN-to-GNN Knowledge Distillation.}
Despite the great progress, most existing GNNs share the de facto design that relies on message passing to aggregate features from neighborhoods, which may be one major source of latency in GNN inference. To address this problem, there are previous works that attempt to distill knowledge from large teacher GNNs to smaller student GNNs, termed as GNN-to-GNN distillation \cite{lassance2020deep,zhang2020iterative,ren2021multi,joshi2021representation,wu2022knowledge,wu2022teaching}. For example, the student model in RDD \cite{zhang2020reliable} and TinyGNN \cite{yan2020tinygnn} is a GNN with fewer parameters but not necessarily fewer layers than the teacher GNN. Besides, LSP \cite{yang2020distilling} transfers the topological structure (rather than feature) knowledge from a pre-trained teacher GNN to a shallower student GNN. In addition, GNN-SD \cite{chen2020self} directly distills knowledge across different GNN layers, mainly aiming to solve the over-smoothing problem but with unobvious performance improvement at shallow layers. Moreover, FreeKD \cite{feng2022freekd} studies a free-direction knowledge distillation architecture, with the purpose of dynamically exchanging knowledge between two shallower GNNs. Note that both teacher and student models in the above works are GNNs, making it still suffer from neighborhood-fetching latency.

\noindent \textbf{GNN-to-MLP Knowledge Distillation.}
To enjoy the topology awareness of GNNs and inference-efficient of MLPs, the other branch of graph knowledge distillation is to directly distill from teacher GNNs to lightweight student MLPs, termed as GNN-to-MLP distillation. For example, CPF \cite{yang2021extract} \textit{directly} improves the performance of student MLPs by adopting deeper/wider network architectures and incorporating label propagation in MLPs, both of which burden the inference latency. Instead, GLNN \cite{zhang2021graph} distills knowledge from teacher GNNs to vanilla MLPs without other computing-consuming operations; while the performance of their distilled MLPs can be \textit{indirectly} improved by employing more powerful GNNs, they still cannot match GNN-to-GNN distillation in terms of classification performance. To further improve GLNN, RKD-MLP \cite{anonymous2023double}  adopts a meta-policy to filter out unreliable soft labels, but this is essentially a down-sampling-style strategy that will further reduce the already limited supervision. In contrast, this paper aims to provide more reliable supervision for training student MLPs, which can be considered as an up-sampling-style strategy.

\vspace{-0.6em}
\section{Preliminaries}
\vspace{-0.3em}
\textbf{Notions and Problem Statement.} 
Let $\mathcal{G}=(\mathbf{A}, \mathbf{X})$ be a graph with the node set $\mathcal{V}$ and edge set $\mathcal{E}$, where $\mathcal{V}$ is the set of $N$ nodes with features $\mathbf{X}=\left[\mathbf{x}_{1}, \mathbf{x}_{2}, \cdots, \mathbf{x}_{N}\right]\in \mathbb{R}^{N \times d}$. The graph structure is denoted by an adjacency matrix $\mathbf{A} \in[0,1]^{N \times N}$ with $\mathbf{A}_{i,j}=1$ if $e_{i,j}\in\mathcal{E}$ and $\mathbf{A}_{i,j}=0$ if $e_{i,j} \notin \mathcal{E}$. Considering a semi-supervised node classification task where only a subset of node $\mathcal{V}_L$ with labels $\mathcal{Y}_L$ are known, we denote the labeled set as $\mathcal{D}_L=(\mathcal{V}_L,\mathcal{Y}_L)$ and unlabeled set as $\mathcal{D}_U=(\mathcal{V}_U,\mathcal{Y}_U)$, where $\mathcal{V}_U=\mathcal{V} \backslash \mathcal{V}_L$. The node classification aims to learn a mapping $\Phi: \mathcal{V} \rightarrow \mathcal{Y}$ so that it can be used to infer the ground-truth label $y_i\in\mathcal{Y}_U$.

\begin{figure*}[!tbp]
	\begin{center}
		\subfigure[Confidence Distribution]{\includegraphics[width=0.3\linewidth]{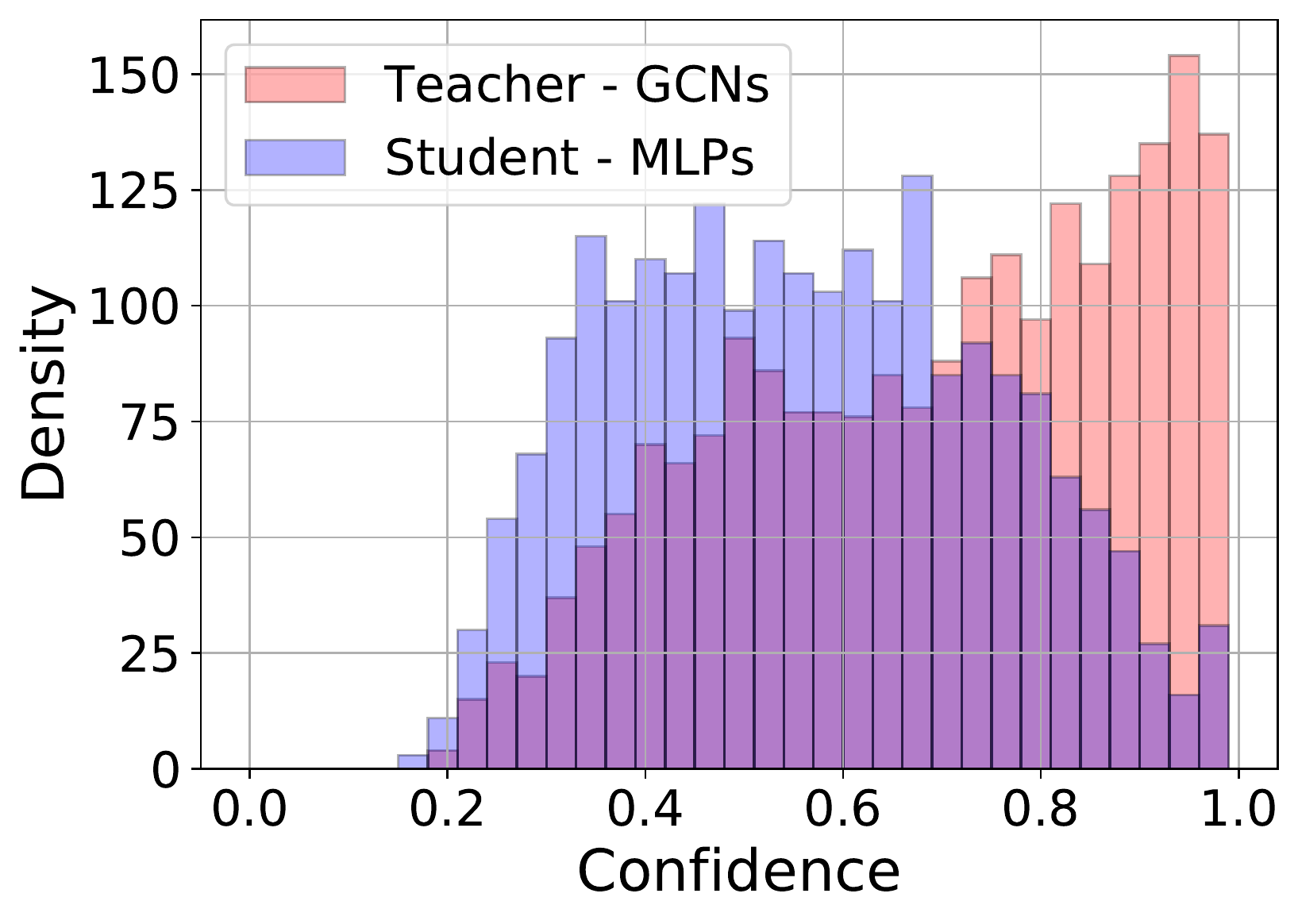}\label{fig:2a}}
		\subfigure[Histogram of \emph{False Negative} Samples]{\includegraphics[width=0.3\linewidth]{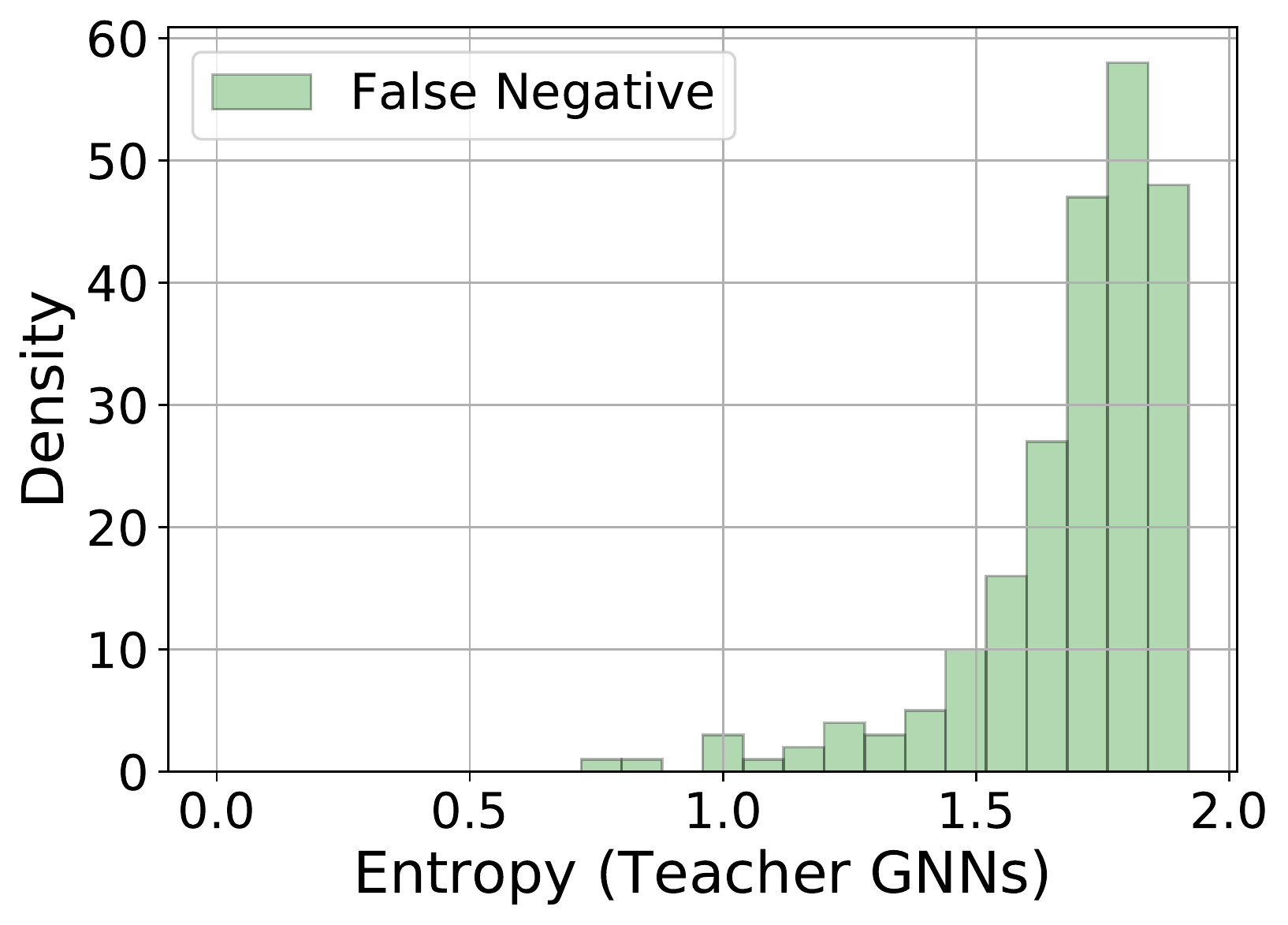}\label{fig:2b}}
		\subfigure[GNN Entropy \textit{vs.} MLP Confidence]{\includegraphics[width=0.3\linewidth]{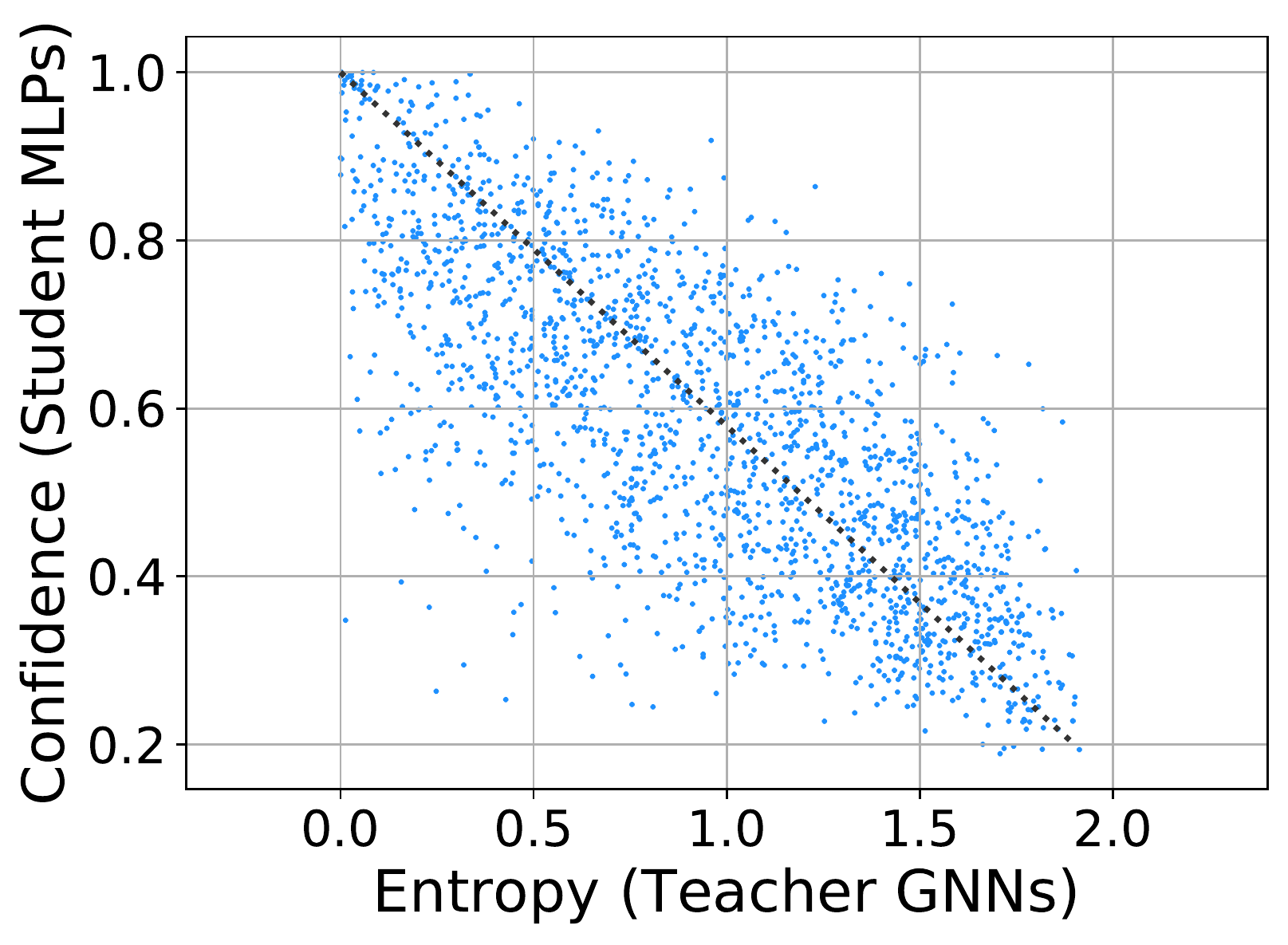}\label{fig:2c}}
	\end{center}
	\vspace{-1em}
	\caption{\textbf{\emph{(a)}} Histograms of the confidence distributions of teacher GCNs and students MLP for those correct predictions on the \texttt{Cora} dataset. \textbf{\emph{(b)}} Distribution of \emph{``False Negative"} samples w.r.t the information entropy of teacher's predictions on the \texttt{Cora} dataset. \textbf{\emph{(c)}} Scatter curve of confidence (student MLP) and information entropy (teacher GCN) for those \emph{``True Positive"} samples on the \texttt{Cora} dataset.}
	\vspace{-1em}
	\label{fig:2}
\end{figure*}

\noindent\textbf{Graph Neural Networks (GNNs).}
A general GNN framework consists of two key operations for each node $v_i$: (1) $\operatorname{AGGREGATE}$: aggregating messages from neighborhood $\mathcal{N}_i$; (2) $\operatorname{UPDATE}$: updating node representations. For an $L$-layer GNN, the formulation of the $l$-th layer is as
\vspace{-0.3em}
\begin{equation}
\begin{small}
\begin{aligned}
\mathbf{m}_{i}^{(l)}  =& \operatorname{AGGREGATE}^{(l)}\left(\big\{\mathbf{h}_{j}^{(l-1)}: v_{j} \in \mathcal{N}_i\big\}\right) \\ 
\mathbf{h}_{i}^{(l)}  =& \operatorname{UPDATE}^{(l)}\left(\mathbf{h}_{i}^{(l-1)}, \mathbf{m}_{i}^{(l)}\right)
\label{equ:1}
\end{aligned}
\end{small}
\vspace{-0.9em}
\end{equation}
where $1\leq l \leq L$, $\mathbf{h}_{i}^{(0)}=\mathbf{x}_{i}$ is the input node feature, and $\mathbf{h}_{i}^{(l)}$ is the node representation of node $v_i$ in the $l$-th layer.

\noindent\textbf{Multi-Layer Perceptrons (MLPs).}
To achieve efficient inference, the vanilla MLPs are used as the student model by default in this paper. For a $L$-layer MLP, the $l$-th layer is composed of a linear transformation, an activation function $\mathrm{ReLu}(\cdot)$, and a dropout function $\operatorname{Dropout}(\cdot)$, as follows
\vspace{-0.3em}
\begin{equation}
\mathbf{z}^{(l)}_i=\operatorname{Dropout}\big(\mathrm{ReLu}\big(\mathbf{z}^{(l-1)}_i \mathbf{W}^{(l-1)}\big)\big) 
\label{equ:2}
\vspace{-0.3em}
\end{equation}
where $\mathbf{z}^{(0)}_i=\mathbf{x}_i$ is the input feature, and $\{\mathbf{W}^{(l)}\}_{l=0}^{L-1}$ are weight matrices with the hidden dimension $F$. In this paper, the network architecture of MLPs, such as the layer number $L$ and layer size $F$, is set the same as that of teacher GNNs.
\newline
\vspace{-1em}

\noindent\textbf{GNN-to-MLP Knowledge Distillation.}
The knowledge distillation is first introduced in \cite{hinton2015distilling} to mainly handle image data. However, recent works on GNN-to-MLP distillation \cite{yang2021extract,zhang2021graph} extend it to the graph domain by imposing KL-divergence constraint $\mathcal{D}_{KL}(\cdot, \cdot)$ between the label distributions generated by teacher GNNs and student MLPs, as follows
\vspace{-0.3em}
\begin{equation}
\begin{small}
\begin{aligned}
\mathcal{L}_{\mathrm{KD}}=\frac{1}{|\mathcal{V}|}\sum_{i\in\mathcal{V}}\mathcal{D}_{KL}\left(\sigma\big(\mathbf{z}_{i}^{(L)}\big), \sigma\big(\mathbf{h}_{i}^{(L)}\big)\right)
\label{equ:3}
\end{aligned}
\end{small}
\vspace{-0.7em}
\end{equation}
where $\sigma(\cdot)=\operatorname{softmax}(\cdot)$, and all nodes (knowledge points) in the set $\mathcal{V}$ are indiscriminately used as supervisions.

\vspace{-0.7em}
\section{Methodology}
\vspace{-0.3em}
\subsection{What Gets in the Way of Better Distillation?}
\begin{figure*}[!tbp]
	\begin{center}
		\subfigure[Visualizations in GNNs]{\includegraphics[width=0.245\linewidth]{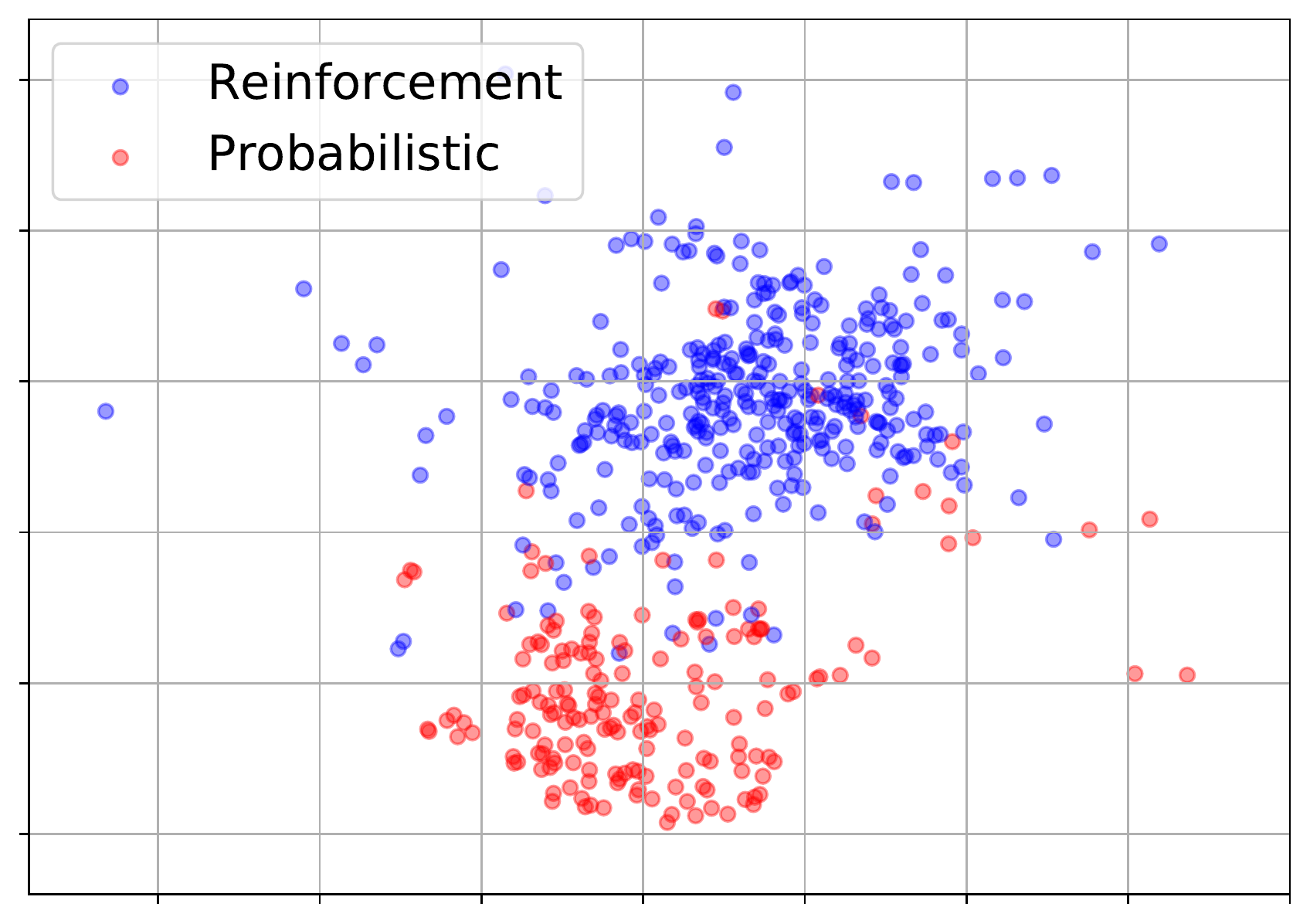}\label{fig:3a}}
        \subfigure[Spatial Distribution in GNNs]{\includegraphics[width=0.245\linewidth]{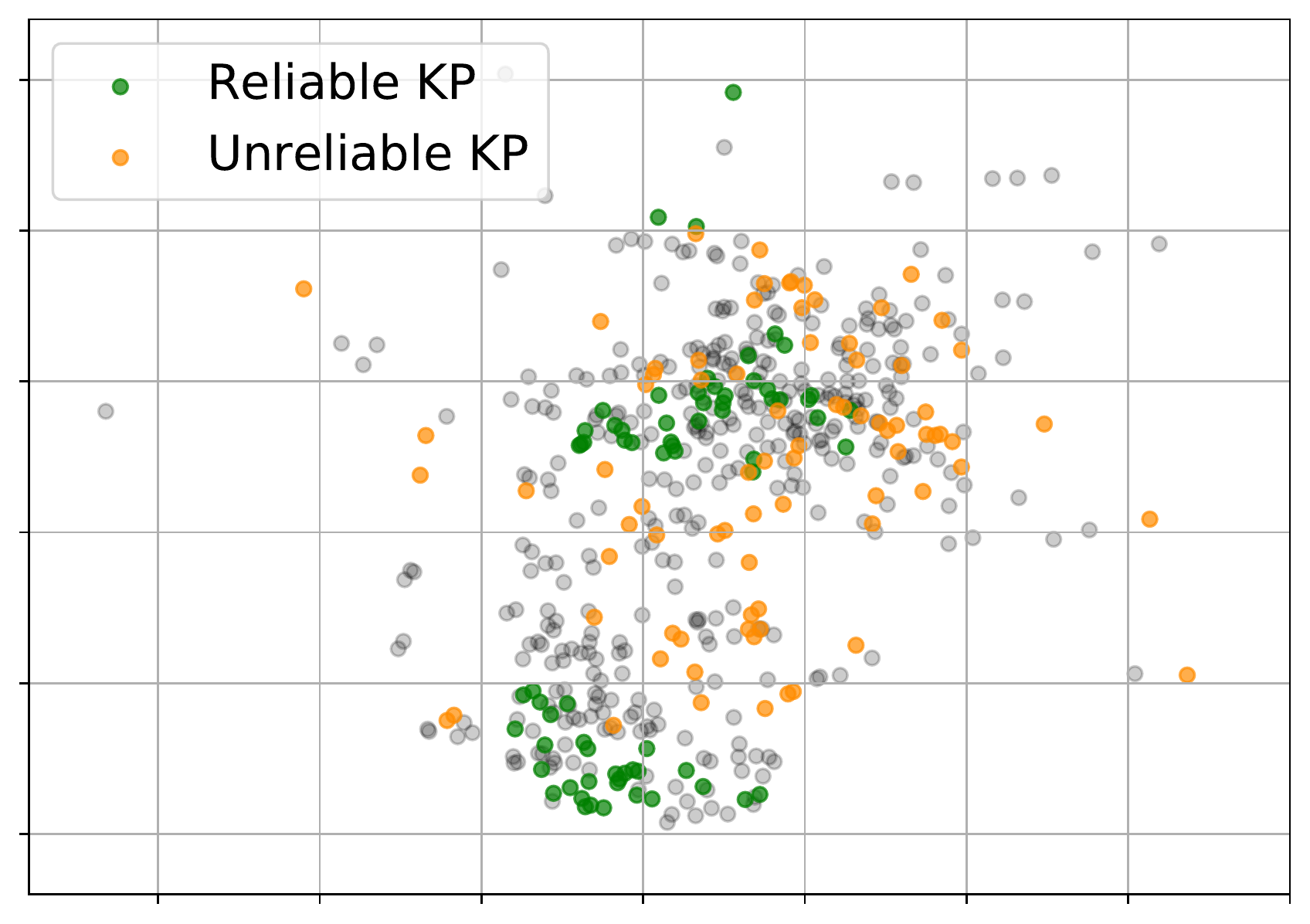}\label{fig:3b}}
		\subfigure[Visualizations in MLPs]{\includegraphics[width=0.245\linewidth]{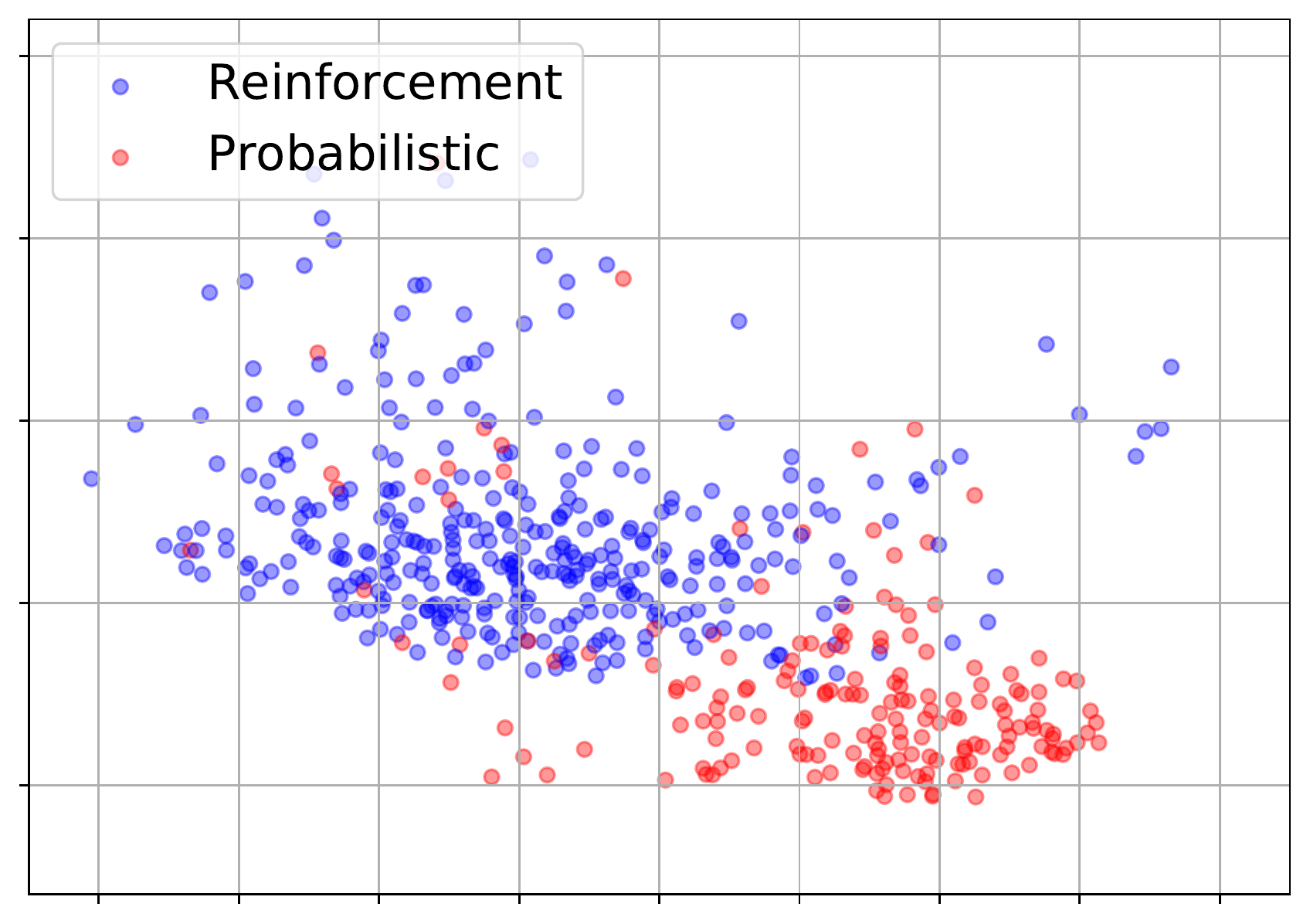}\label{fig:3c}}
		\subfigure[Spatial Distribution in MLPs]{\includegraphics[width=0.245\linewidth]{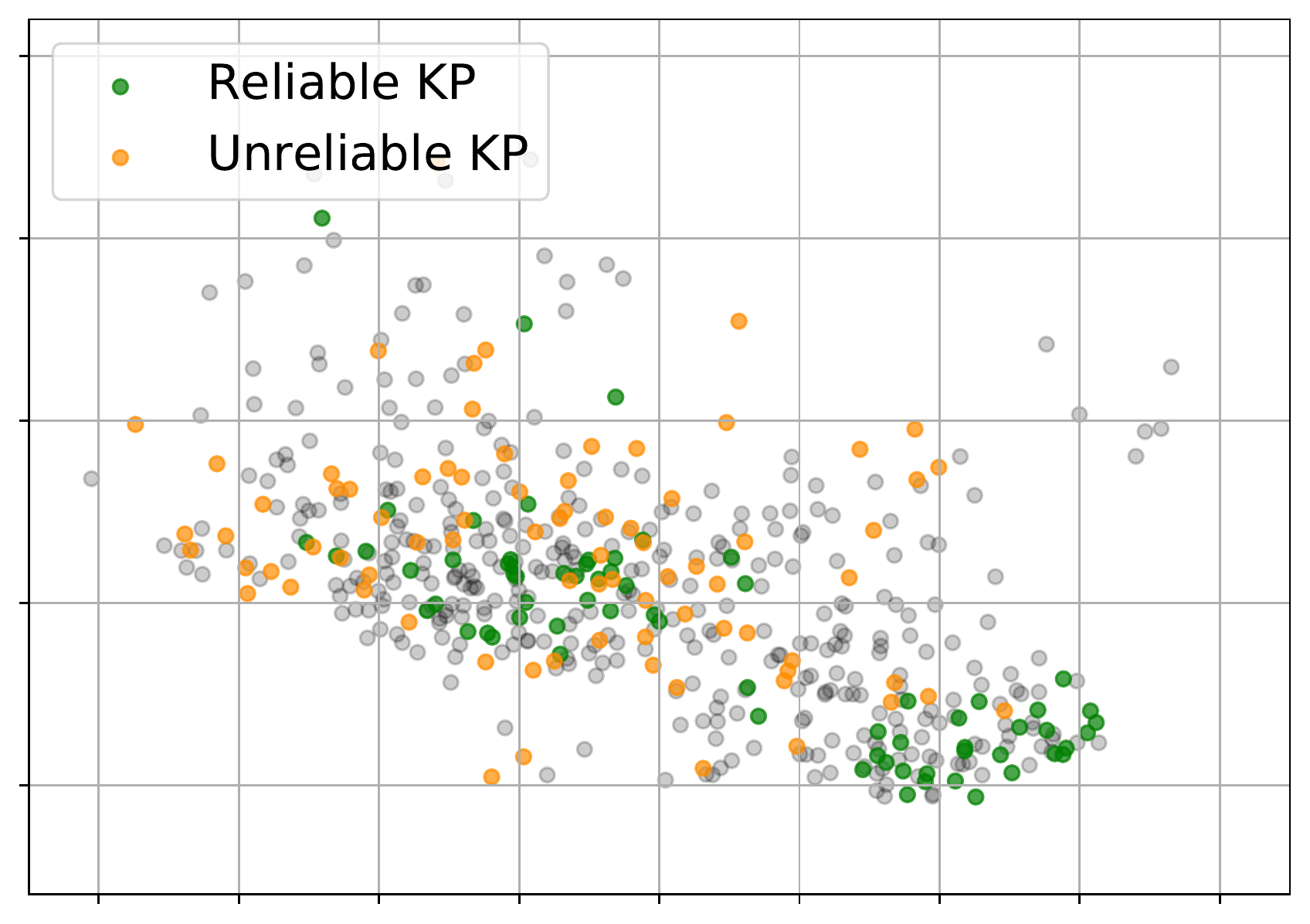}\label{fig:3d}}
	\end{center}
	\vspace{-1em}
	\caption{\textbf{\emph{(a)(c)}} Visualizations of the embeddings of teacher GNNs and student MLPs for two classes on \texttt{Cora}. \textbf{\emph{(b)(d)}} Spatial distribution of knowledge points with the reliability ranked in the top 20\% and bottom 10\%, which are marked in {\color[rgb]{0.2156,0.7176,0.2773}green} and {\color[rgb]{1.0,0.6,0.0}orange}, respectively.}
	\vspace{-1em}
	\label{fig:3}
\end{figure*}

\noindent\textbf{Potential Under-confident Problem}. The GNN-to-MLP distillation can be achieved by directly optimizing the objective function $\mathcal{L}_{\mathrm{KD}}$ defined in Eq.~(\ref{equ:3}). However, such a straightforward distillation completely ignores the differences between knowledge points in GNNs and may suffer from a potential under-confident problem, i.e., the distilled MLP may fail to make predictions as confidently as teacher GNNs. To illustrate this problem, we report in Fig.~\ref{fig:2a} the confidences of teacher GCNs and student MLPs for those correct predictions by the UMAP \cite{mcinnes2018umap} algorithm on the Cora dataset. It can be seen that there exists a significant distribution shift between the confidence distribution of teacher GCNs and student MLPs, which confirms the existence of the under-confident problem. The direct hazard of such an under-confident problem is that it may push those samples located near the class boundaries into incorrect predictions, as shown in Fig.~\ref{fig:3a} and Fig.~\ref{fig:3c}, which hinders the performance of student MLPs.

To go deeper into the under-confident problem and explore what exactly stands in the way of better GNN-to-MLP distillation, we conducted extensive theoretical and experimental analysis and found that one of the main causes could be due to the lack of reliable supervision from teacher GNNs.

\noindent\textbf{Theoretical Analysis}. The main strength of teacher GNNs over student MLPs is their excellent topology-awareness capability, which is mainly enabled by message passing. There have been a number of works exploring the roles of message passing in GNNs. For example, \cite{yang2020revisiting} have proved that message passing (architecture design) in GNNs is equivalent to performing Laplacian smoothing (supervision design) on node embeddings in MLPs. In essence, message-passing-based GNNs implicitly take the objective of Dirichlet energy minimization \cite{belkin2001laplacian} as graph-based regularization, which is defined as follows
\vspace{-0.3em}
\begin{equation}
\mathcal{L}_{reg}= \operatorname{Tr}\left(\mathbf{Y}^{\top} \Delta \mathbf{Y}\right) = \sum_{i} \sum_{j\in\mathcal{N}_i}\Big\|\frac{\mathbf{Y}_i}{\sqrt{d_i}}-\frac{\mathbf{Y}_j}{\sqrt{d_j}}\Big\|^2_2
\vspace{-0.6em}
\end{equation}
where $\Delta=\mathbf{I}-\mathbf{D}^{-\frac{1}{2}} \mathbf{A} \mathbf{D}^{-\frac{1}{2}}$ is the normalized Laplacian operator, $\mathbf{D}$ is the degree matrix with $\mathbf{D}_{i,i}=d_i=\sum_{j}\mathbf{A}_{i,j}$, and $\mathbf{Y}=\operatorname{softmax}\left(\mathbf{H}^{(L)}\right)$ is the label distribution matrix.

Apart from the supervision of cross-entropy on the labeled set, message passing in GNNs implicitly provides a special kind of \textbf{\emph{self-supervision}}, which imposes regularization constraints on the label distributions between neighboring nodes. We conjecture that it is exactly such additional self-supervision that enables GNNs to make highly confident predictions.
In contrast, student MLPs are trained in a way that cannot capture the fine-grained dependencies between neighboring nodes; instead, they only learn the overall contextual information about their neighborhood from teacher GNNs, resulting in undesirable under-confident predictions.

\vspace{-0.2em}
\noindent\textbf{Experimental Analysis}. To see why the (distilled) student MLPs tend to make low-confidence predictions, we conducted an in-depth statistical analysis on two types of special samples. \emph{(1)} The distribution of \emph{``False Negative"} samples (predicted correctly by GNNs but incorrectly by MLPs) w.r.t the information entropy of teacher's predictions is reported in Fig.~\ref{fig:2b}, from which we observe that most of the \emph{``False Negative"} samples are distributed in the region of higher entropy. \emph{(2)} For those \emph{``True Positive"} samples (predicted correctly by both GNNs and MLPs), the scatter of confidence and information entropy from student MLPs and teacher GNNs is plotted in Fig.~\ref{fig:2c}, which shows that GNN knowledge with high uncertainty (low reliability) may undermine the capability of student MLPs to make sufficiently confident predictions. Based on these two observations, it is reasonable to hypothesize that one cause of the under-confident problem suffered by student MLPs is be the lack of sufficiently reliable supervision from teacher GNNs.

\vspace{-0.7em}
\subsection{How to Quantify the Knowledge in GNNs?} \label{sec:4.2}
\vspace{-0.3em}
Based on the above experimental and theoretical analysis, a key issue in GNN-to-MLP distillation may be to provide \textbf{\emph{more and reliable}} supervision for training student MLPs. Next, we first describe how to quantify the reliability of knowledge in GNNs, and then propose how to sample more reliable supervision through a knowledge-inspired manner.

\noindent\textbf{Knowledge Quantification.} Given a graph $\mathcal{G} = (\mathbf{A}, \mathbf{X})$ and a pre-trained teacher GNN $f_\theta(\cdot,\cdot)$, we propose to quantify the reliability of a knowledge point (node) $v_i\in\mathcal{V}$ in GNNs by measuring the invariance of its information entropy to noise perturbations, which is defined as follows
\vspace{-0.3em}
\begin{equation}
\begin{small}
\begin{aligned}
    \rho_i = \frac{1}{\delta^2} \underset{\mathbf{X}^{\prime} \sim \mathcal{N}(\mathbf{X}, \boldsymbol{\Sigma}(\delta))}{\mathbb{E}}\left[\left\|\mathcal{H}(\mathbf{Y}^{\prime}_i)-\mathcal{H}(\mathbf{Y}_i)\right\|^2\right],
    \\ \text{where} \quad \mathbf{Y}^{\prime}=f_\theta(\mathbf{A}, \mathbf{X}^{\prime}) \ \ \ \text{and}\ \ \  \mathbf{Y}=f_\theta(\mathbf{A}, \mathbf{X})
\end{aligned}
\end{small}
\label{equ:5}
\end{equation}
where $\delta$ is the variance of Gaussian noise and $\mathcal{H}(\cdot)$ denote the information entropy. The smaller the metric $\rho_i$ is, the higher the reliability of knowledge point $v_i$ is. The quantification of GNN knowledge defined in Eq.~(\ref{equ:5}) has the following three strengths: \textbf{\emph{(1)}} It measures the robustness of knowledge in teacher GNNs to noise perturbations, and thus more truly reflects the reliability of different knowledge points, which is very important for reliable distillation. \textbf{\emph{(2)}} The message passing is what makes GNNs special over MLPs, so the key to quantify GNN knowledge is to measure its topology-awareness capability. Compared with node-wise information entropy, Eq.~(\ref{equ:5}) not only reflects the node uncertainty, but also takes into account the contextual information from the neighborhood. \textbf{\emph{(3)}} As will be analyzed next, the knowledge quantified by Eq.~(\ref{equ:5}) shows the roles played by different knowledge points \emph{spatially} and \emph{temporally}. 

\noindent\textbf{Spatial Distribution of Knowledge Points}. To explore the spatial distribution of different knowledge points in the graph, we first visualize the embeddings of teacher GNNs and student MLPs in Fig.~\ref{fig:3a} and Fig.~\ref{fig:3c}, and then we mark the knowledge points with the reliability ranked in the top 20\% and bottom 10\% as {\color[rgb]{0.2156,0.7176,0.2773}green} and {\color[rgb]{1.0,0.6,0.0}orange} in Fig.~\ref{fig:3b} and Fig.~\ref{fig:3d}. To make it clearer, we only report the results for two classes on the Cora dataset; more visualizations can be found in \textbf{Appendix C}. We find that different knowledge points are differentially distributed in the graph, where most reliable knowledge points are distributed around the class centers regardless of being in teacher GNNs or student MLPs, while those unreliable ones are distributed at the class boundaries. The spatial distribution of knowledge points explains well why most of the \emph{False Negatice} samples are located in regions with high uncertainty in Fig.~\ref{fig:2c}.

\vspace{-0.5em}
\begin{figure}[!htbp]
	\begin{center}
		\includegraphics[width=0.75\linewidth]{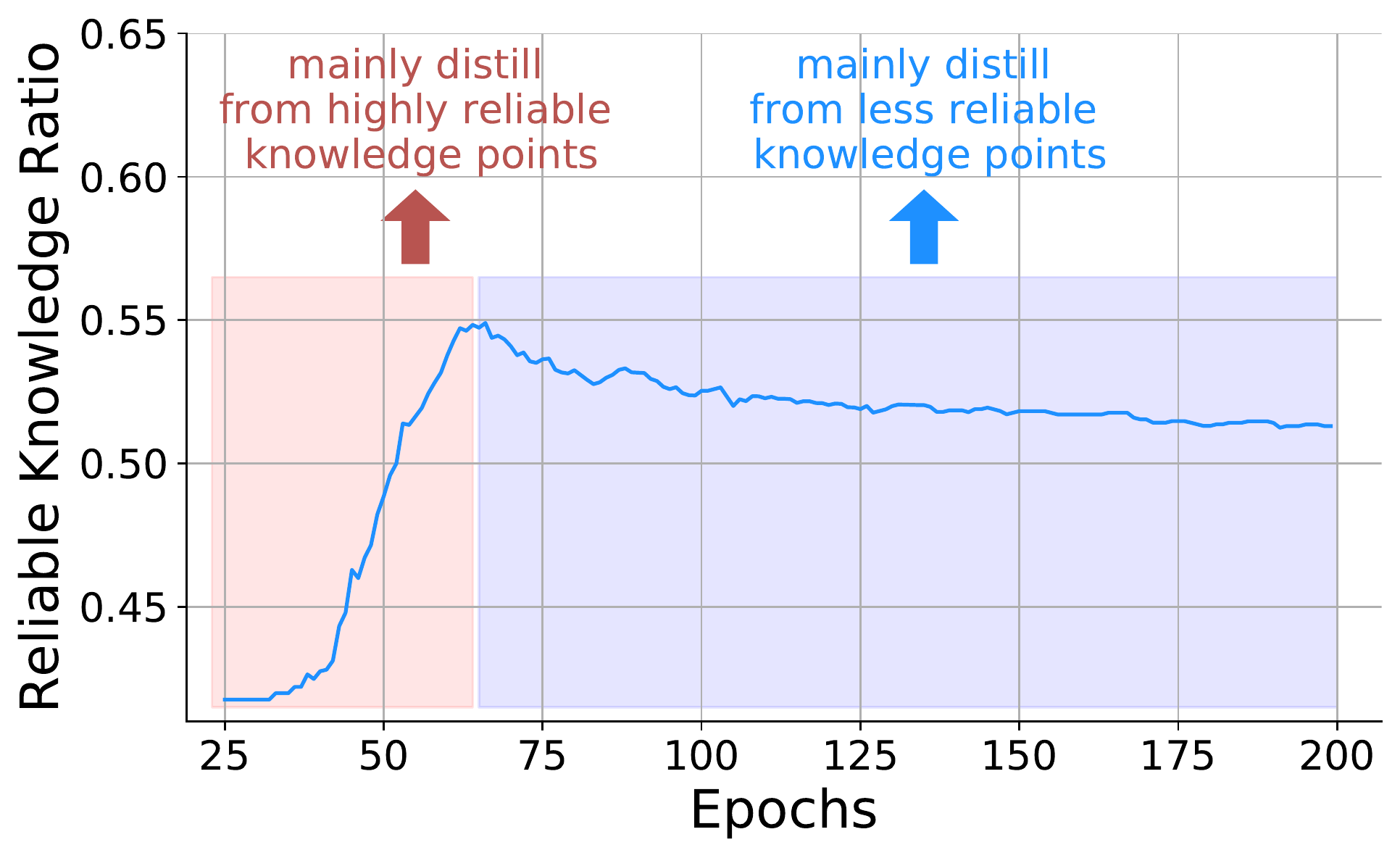}
	\end{center}
	\vspace{-1.5em}
	\caption{Percentage of highly reliable knowledge points on \texttt{Cora} to show the distillation speeds of different knowledge points.}
	\vspace{-0.5em}
	\label{fig:4}
\end{figure}

\noindent\textbf{Temporal Distribution of Knowledge Points}. To see the distillation speed of different knowledge points, we explore which knowledge points the student MLPs will be fitted to first during the training process. We considered those knowledge points that are correctly predicted by student MLPs and ranked in the top 50\% of reliability, among which we calculate the percentage of points with the top 20\% of reliability in Fig.~\ref{fig:4}. It can be seen that student MLPs will quickly fit to those highly reliable knowledge points first as the training proceeds, and then gradually learn from those relatively less reliable knowledge points. This indicates that different knowledge points play different roles in the distillation process, which inspires us to sample some reliable knowledge points from teacher GNNs \emph{in a dynamic manner} to provide more additional supervision for training MLPs.

\subsection{Knowledge-inspired Reliable Distillation}
\begin{figure}[!htbp]
	\begin{center}
		\includegraphics[width=1.0\linewidth]{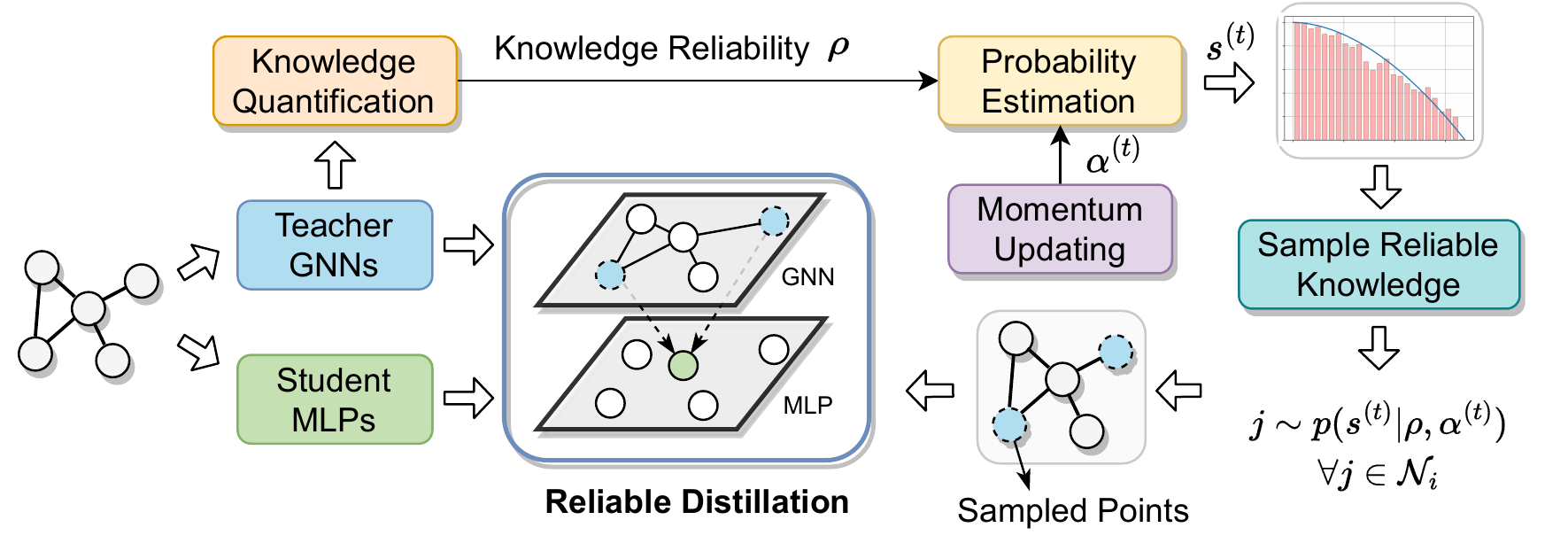}
	\end{center}
	\vspace{-1.5em}
	\caption{A high-level overview of the proposed \texttt{KRD} framework.}
	\vspace{-0.3em}
	\label{fig:5}
\end{figure}

In this subsection, we first model the probability of each node being an informative and reliable knowledge point based on the knowledge quantification defined by Eq.~(\ref{equ:5}). Next, we propose a knowledge-based sampling strategy to make full use of those reliable knowledge points as additional supervision for more reliable distillation into MLPs. A high-level overview of the proposed \emph{Knowledge-inspired Reliable Distillation} (\texttt{KRD}) framework is shown in Fig.~\ref{fig:5}.

\vspace{-0.5em}

\begin{figure}[!htbp]
	\begin{center}
    \includegraphics[width=0.49\linewidth]{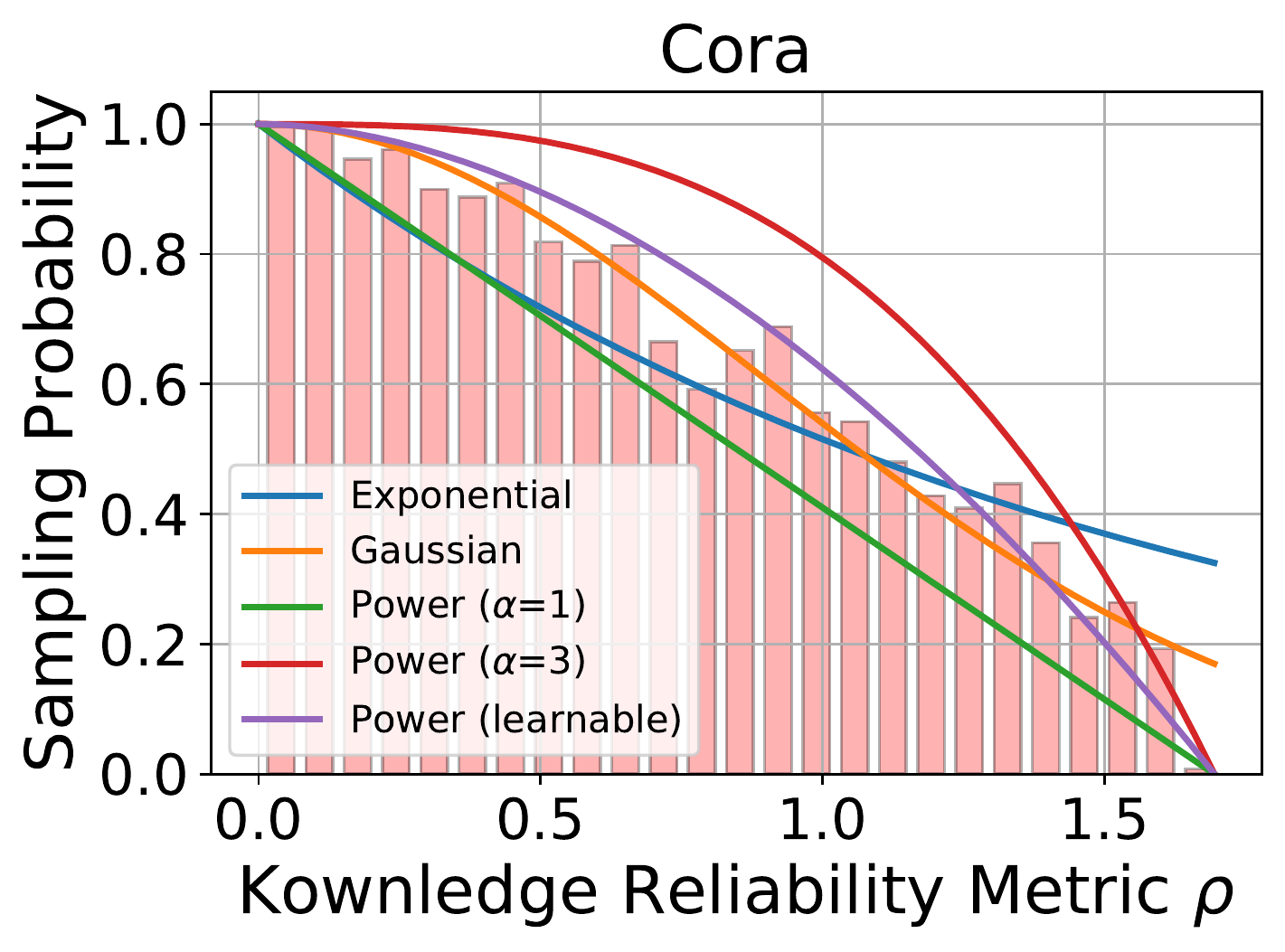}
    \includegraphics[width=0.49\linewidth]{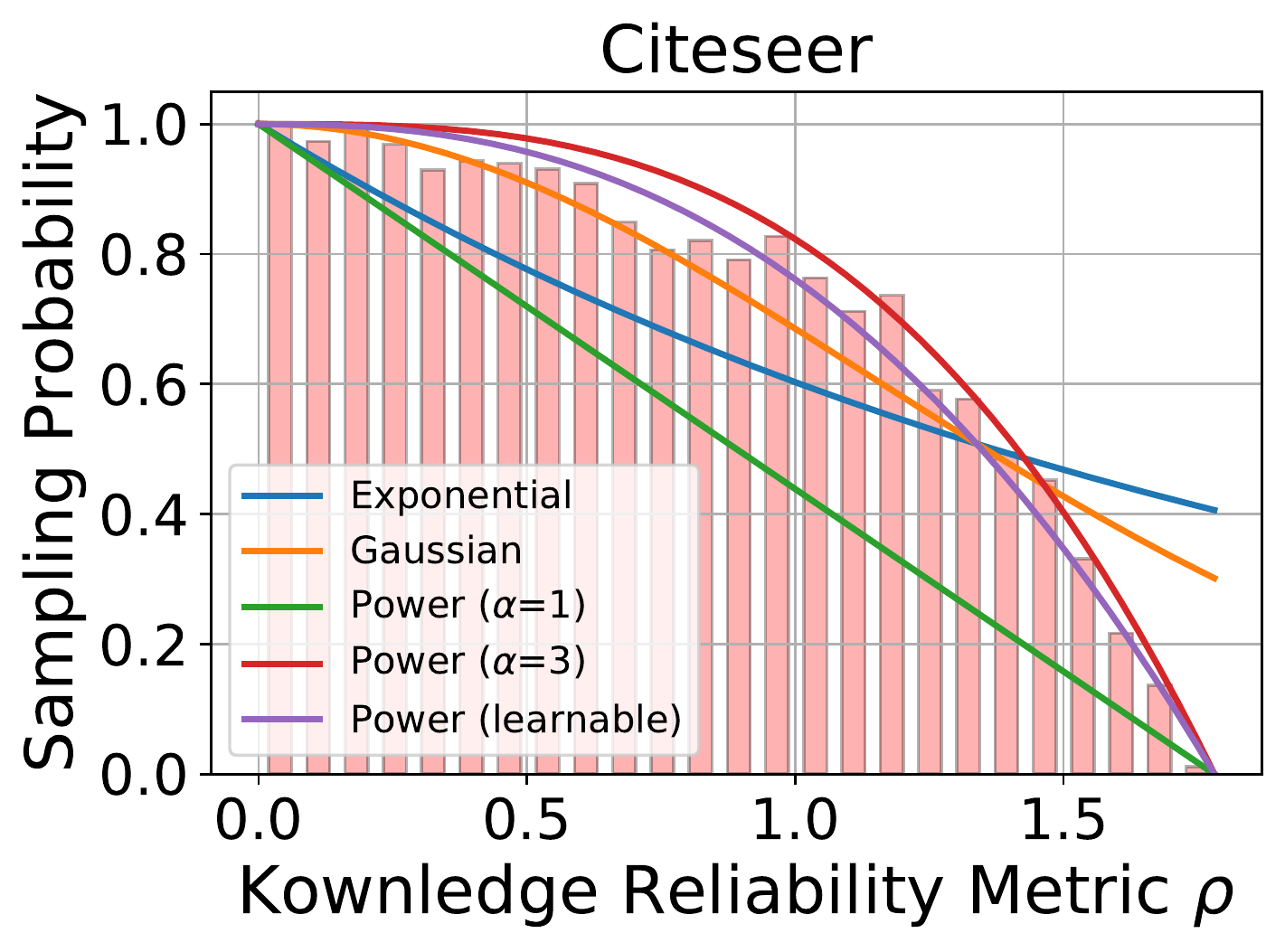}
	\end{center}
	\vspace{-1em}
	\caption{Histograms of \emph{``True Positive"} sample density w.r.t the reliability metric $\rho$, as well as five distribution fitting schemes for modeling the sampling probability on the \texttt{Cora} and \texttt{Citeseer} datasets, where the density has been min/max normalized.}
	\vspace{-0.5em}
	\label{fig:6}
\end{figure}

\noindent\textbf{Sampling Probability Modeling}. We aim to estimate the sampling probability of a knowledge point based on its quantified reliability. As shown in Fig.~\ref{fig:6}, we plot the histograms of \emph{``True Positive"} sample density w.r.t the reliability metric $\rho$ on two datasets (see \textbf{Appendix D} for more results), where the density has been min/max normalized. We model the sampling probability $s_i$ of node $v_i$ based on the metric $\rho_i$ by a learnable
power distribution (with power $\alpha$), as follows:
\vspace{-0.5em}
\begin{equation}
    p(s_i \mid \rho_i, \alpha) = 1 - (\frac{\rho_i}{\rho_M})^{\alpha}, \quad \forall v_i \in \mathcal{V}
\label{equ:6}
\vspace{-0.5em}
\end{equation}
where $\rho_M=\operatorname{argmax}_j \rho_j$. When the ground-truth labels are available, an optimal power $\alpha_{opt}$ can be directly fitted from histograms. However, the ground-truth labels are often unknown in practice, so we propose to combine the student MLPs $g_{\psi^{(t)}}(\cdot)$ with the pre-trained teacher GNNs $f_{\theta_{pre}}(\cdot)$ to model $p\left(\alpha^{(t)} \mid f_{\theta_{pre}}(\mathbf{A}, \mathbf{X}), g_{\psi^{(t)}}(\mathbf{A}, \mathbf{X})\right)$ at $t$-th epoch, which can be implementated by the following four steps: (1) initializing the power $\alpha^{(0)}=1.0$; (2) constructing a histogram of sample density (predicted to be the same by both teacher GNNs and student MLPs) w.r.t the knowledge reliability metric $\rho$; (3) inferring a new power $\alpha^{(t)}_{new}$ by fitting the histogram; (4) updating power $\alpha^{(t-1)}$ in a dynamic momentum manner, which can be formulated as follows
\begin{equation}
    \alpha^{(t)} \leftarrow \eta\alpha^{(t-1)} + (1-\eta) * \alpha^{(t)}_{new}
    \label{equ:7}
\end{equation} 
where $\eta$ is the momentum updating rate. We provide the fitted curves with \emph{fixed} and \emph{learnable} powers in Fig.~\ref{fig:6}, which shows that the fitted distributions of learnable powers are more in line with the histogram. Moreover, we also include the results of fitting by Gaussian and exponential distributions as comparisons, but it shows that they do not work better. A quantitative comparison of different distribution fitting schemes has been provided in Table.~\ref{tab:2}, and the fitted results on more datasets are available in \textbf{Appendix D}.

\vspace{-0.5em}

\noindent\textbf{Knowledge-based Sampling}. Next, we describe how to sample a set of reliable knowledge points as additional supervision for training student MLPs. Given any target node $v_i$, we first sample some highly reliable knowledge points $v_j \in \mathcal{N}_i$ from its neighborhood according to the sampling probability $p(s_j\mid\rho_i, \alpha^{(t)})$. Then, we take sampled knowledge points as multiple teachers and distill their knowledge into student MLPs as additional supervision through a \emph{multi-teacher distillation} objective, which is defined as follows
\vspace{-0.5em}
\begin{equation}
\begin{small}
\begin{aligned}
\mathcal{L}_{\mathrm{KRD}}\!=\!\underset{i}{\mathbb{E}}\underset{j\sim p(s_j\mid\rho_i, \alpha^{(t)})}{\underset{j\in\mathcal{N}_i}{\mathbb{E}}} 
\mathcal{D}_{KL}\Big(\sigma(\mathbf{z}_{j}^{(L)} / \tau), \sigma(\mathbf{h}_{i}^{(L)} / \tau)\Big)
\end{aligned}
\end{small}
\label{equ:8}
\end{equation}
where $\tau$ is the distillation temperature coefficient.

\vspace{-0.5em}
\subsection{Training Strategy} 
The pseudo-code of the \texttt{KRD} framework is summarized in Algorithm~\ref{algo:1}. To achieve GNN-to-MLP knowledge distillation, we first pre-train the teacher GNNs with the classification loss  $\mathcal{L}_{\mathrm{label}}=\frac{1}{|\mathcal{V}_L|}\sum_{i\in\mathcal{V}_L}\operatorname{CE}\big(y_i, \sigma(\mathbf{h}_i^{(L)})\big)$, where $\operatorname{CE}(\cdot)$ denotes the cross-entropy loss. Finally, the total objective function to distill reliable knowledge from the teacher GNNs into the student MLPs is defined as follows
\begin{equation*}
\mathcal{L}_{\mathrm{total}} =  \frac{\lambda}{|\mathcal{V}_L|}\sum_{i\in\mathcal{V}_L}\mathcal{H}\big(y_i, \sigma(\mathbf{z}_i^{(L)})\big) + \big(1-\lambda\big) \big(\mathcal{L}_{\mathrm{KD}} + \mathcal{L}_{\mathrm{KRD}}\big)
\end{equation*}
where $\lambda$ is the weight to balance the influence of the classification loss and knowledge distillation losses.

\vspace{-0.5em}
\subsection{Time Complexity Analysis}
It is noteworthy that the main computational burden introduced in this paper comes from additional reliable supervision as defined in Eq.~(\ref{equ:8}). However, we sample reliable knowledge points \emph{in the neighborhood} instead of the entire set of nodes $\mathcal{V}$, which has reduced the time complexity from $\mathcal{O}(|\mathcal{V}^2|F)$ to less than $\mathcal{O}(|\mathcal{E}|F)$. The training time complexity of the \texttt{KRD} framework mainly comes from two parts: (1) GNN training $\mathcal{O}(|\mathcal{V}|dF + |\mathcal{E}|F)$ and (2) knowledge distillation $\mathcal{O}(|\mathcal{E}|F)$, where $d$ and $F$ are the dimensions of input and hidden spaces. The total time complexity $\mathcal{O}(|\mathcal{V}|dF + |\mathcal{E}|F)$ is linear w.r.t the number of nodes $|\mathcal{V}|$ and edges $|\mathcal{E}|$, which is in the same order as GCNs and GLNN.

\begin{algorithm}[!htbp]
    \caption{Algorithm for KRD framework (Transductive)}
    \label{algo:1}
    \begin{algorithmic}[1]
  
    \REQUIRE  Graph $\mathcal{G}=(\mathcal{V},\mathcal{E})$, Node Features: $\mathbf{X}$, \# Epoch: $E$.
    \ENSURE Predicted labels $\mathcal{Y}_U$, MLP parameters $\{\mathbf{W}^{l}\}_{l=0}^{L-1}$.
  
    \STATE Randomly initialize the parameters of GNNs and MLPs.
	\STATE Pre-train the teacher GNNs until convergence by $\mathcal{L}_{\mathrm{label}}$.
	\STATE Quantify the reliability of knowledge points by Eq.~(\ref{equ:5}).
  
    \FOR{$t$ $\in$ \{1, 2, $\cdots$, $E+1$\}}
        \STATE Calculate the node representations from teacher GNNs and student MLPs by Eq.~(\ref{equ:1}) and Eq.~(\ref{equ:2});
		\STATE Estimate the sampling probability by Eq.~(\ref{equ:6});
		\STATE Sample reliable knowledge points and calculate the multi-teacher distillation loss $\mathcal{L}_{\mathrm{KRD}}$ by Eq.~(\ref{equ:8});
		
		\STATE Calculate the total loss $\mathcal{L}_{\mathrm{total}}$ and update the parameters of student MLPs $\{\mathbf{W}^{l}\}_{l=0}^{L-1}$ by back propagation.
		\STATE Momentum updating the power $\alpha^{(t)}$ by Eq.~(\ref{equ:7}).
	\ENDFOR
	
    \STATE Predicted labels $y_i\in\mathcal{Y}_U$ for those unlabeled nodes $\mathcal{V}_U$.
        
	\STATE \textbf{return} Predicted labels $\mathcal{Y}_U$ and Parameters $\{\mathbf{W}^{l}\}_{l=0}^{L-1}$.
    \end{algorithmic}
\end{algorithm}

\section{Experiments} \label{sec:5}
In this section, we evaluate \texttt{KRD} on seven real-world datasets by answering the following six questions. \textbf{Q1}: How effective is \texttt{KRD} in the transductive and inductive settings? Is \texttt{KRD} applicable to different teacher GNNs? \textbf{Q2:} How does \texttt{KRD} compare to other leading baselines on graph knowledge distillation? \textbf{Q3:} What happens if we model the sampling probability using other distribution functions? \textbf{Q4:} How does \texttt{KRD} perform by applying other heuristic knowledge sampling approach? \textbf{Q5:} Can \texttt{KRD} improve the predictive confidence of distilled MLPs? \textbf{Q6:} How do the two key hyperparameters $\lambda$ and $\eta$ influence the performance of \texttt{KRD}?

\textbf{Dataset.} 
The effectiveness of the \texttt{KRD} framework is evaluated on seven real-world datasets, including Cora \cite{sen2008collective}, Citeseer \cite{giles1998citeseer}, Pubmed \cite{mccallum2000automating}, Coauthor-CS, Coauthor-Physics, Amazon-Photo \cite{shchur2018pitfalls}, and ogbn-arxiv \cite{hu2020open}. A statistical overview of datasets is placed in \textbf{Appendix A}. Besides, each set of experiments is run five times with different random seeds, and the average accuracy and standard deviation are reported. Due to space limitations, we defer the implementation details and hyperparameter settings for each dataset to \textbf{Appendix C} and supplementary materials. 

\noindent \textbf{Baselines.}
Three basic components in knowledge distillation are (1) teacher model, (2) student model, and (3) distillation objective. As a model-agnostic framework, \texttt{KRD} can be combined with any teacher GNN architecture. In this paper, we consider three types of teacher GNNs, including GCN \cite{kipf2016semi}, GraphSAGE \cite{hamilton2017inductive}, and GAT \cite{velivckovic2017graph}. Besides, we adopt pure MLPs (with the same layer number
$L$ and size $F$ as teacher GNNs) as the student model for a fair comparison. The focus of this paper is to provide more reliable self-supervision for GNN-to-MLP distillation. Thus, we only take GLNN \cite{zhang2021graph} as an important benchmark to demonstrate the necessity and effectiveness of additional supervision. Besides, we also compare \texttt{KRD} with some state-of-the-art graph distillation baselines in Table.~\ref{tab:2}, including CPF \cite{yang2021extract}, RKD-MLP \cite{anonymous2023double}, FF-G2M \cite{wu2023extracting}, RDD \cite{zhang2020reliable}, TinyGNN \cite{yan2020tinygnn}, LSP \cite{yang2020distilling}, etc.

\begin{table*}[!htbp]
\begin{center}
\caption{Classificatiom accuracy $\pm$ std (\%) on seven real-world datasets in both transductive and inductive settings, where three different GNN architectures (GCN, GraphSAGE, and GAT) have been considered as the teacher models. The best metrics are marked by \textbf{bold}.}
\label{tab:1}
\resizebox{\textwidth}{!}{
\begin{tabular}{cc|ccccccc|c}

\toprule
\textbf{Teacher} & \textbf{Student} & \texttt{Cora} & \texttt{Citeseer} & \texttt{Pubmed} & \texttt{Photo} & \texttt{CS} & \texttt{Physics} & \texttt{ogbn-arxiv} & \textit{Average} \\ \midrule

\multicolumn{10}{c}{\textbf{\texttt{Transductive Setting}}} \\ \midrule
MLPs & - & $59.58_{\pm0.97}$ & $60.32_{\pm0.61}$ & $73.40_{\pm0.68}$ & $78.65_{\pm1.68}$ & $87.82_{\pm0.64}$ & $88.81_{\pm1.08}$ & $54.63_{\pm0.84}$ & - \\ \midrule
\multirow{4}{*}{GCN} & - & $81.70_{\pm0.96}$ & $71.64_{\pm0.34}$ & $79.48_{\pm0.21}$ & $90.63_{\pm1.53}$ & $90.00_{\pm0.58}$ & $92.45_{\pm0.53}$ & $\textbf{71.20}_{\pm0.17}$ & - \\
 & GLNN & $82.20_{\pm0.73}$ & $71.72_{\pm0.30}$ & $80.16_{\pm0.20}$ & $91.42_{\pm1.61}$ & $92.22_{\pm0.72}$ & $93.11_{\pm0.39}$ & $67.76_{\pm0.23}$ & - \\
 & KRD (ours) & $\textbf{84.42}_{\pm0.57}$ & $\textbf{74.86}_{\pm0.58}$ & $\textbf{81.98}_{\pm0.41}$ & $\textbf{92.21}_{\pm1.44}$ & $\textbf{94.08}_{\pm0.34}$ & $\textbf{94.30}_{\pm0.46}$ & $70.92_{\pm0.21}$ & - \\
 & \textit{Improv.} & 2.22 & 3.14 & 1.82 & 0.79 & 1.86 & 1.19 & 3.16 & 2.03 \\ \midrule
\multirow{4}{*}{GraphSAGE} & - & $82.02_{\pm0.94}$ & $71.76_{\pm0.49}$ & $79.36_{\pm0.45}$ & $90.56_{\pm1.69}$ & $89.29_{\pm0.77}$ & $91.97_{\pm0.91}$ & $71.06_{\pm0.27}$ & - \\
 & GLNN & $81.86_{\pm0.88}$ & $71.52_{\pm0.54}$ & $80.32_{\pm0.38}$ & $91.34_{\pm1.46}$ & $92.00_{\pm0.57}$ & $92.82_{\pm0.93}$ & $68.30_{\pm0.19}$ & - \\
 & KRD (ours) & $\textbf{84.60}_{\pm0.76}$ & $\textbf{73.68}_{\pm0.68}$ & $\textbf{81.60}_{\pm0.33}$ & $\textbf{92.12}_{\pm1.50}$ & $\textbf{93.93}_{\pm0.40}$ & $\textbf{94.18}_{\pm0.58}$ & $\textbf{71.50}_{\pm0.25}$ & - \\
 & \textit{Improv.} & 2.74 & 2.16 & 1.28 & 0.78 & 1.93 & 1.36 & 3.20 & 1.92 \\ \midrule
\multirow{4}{*}{GAT} & - & $81.66_{\pm1.04}$ & $70.78_{\pm0.60}$ & $79.88_{\pm0.85}$ & $90.06_{\pm1.38}$ & $90.90_{\pm0.37}$ & $91.97_{\pm0.58}$ & $71.08_{\pm0.19}$ & - \\
 & GLNN & $81.78_{\pm0.75}$ & $70.96_{\pm0.86}$ & $80.48_{\pm0.47}$ & $91.22_{\pm1.45}$ & $92.44_{\pm0.41}$ & $92.70_{\pm0.56}$ & $68.56_{\pm0.22}$ & - \\
 & KRD (ours) & $\textbf{84.12}_{\pm0.39}$ & $\textbf{73.06}_{\pm0.59}$ & $\textbf{82.02}_{\pm0.56}$ & $\textbf{92.13}_{\pm1.48}$ & $\textbf{94.35}_{\pm0.29}$ & $\textbf{94.19}_{\pm0.50}$ & $\textbf{71.45}_{\pm0.26}$ & - \\
 & \textit{Improv.} & 2.34 & 2.10 & 1.54 & 0.91 & 1.91 & 1.49 & 2.89 & 1.88 \\ \midrule

\multicolumn{10}{c}{\textbf{\texttt{Inductive Setting}}} \\ \midrule
MLPs & - & $59.20_{\pm1.26}$ & $60.16_{\pm0.87}$ & $73.26_{\pm0.83}$ & $79.02_{\pm1.42}$ & $87.90_{\pm0.58}$ & $89.10_{\pm0.90}$ & $54.46_{\pm0.52}$ & - \\ \midrule
\multirow{4}{*}{GCN} & - & $\textbf{79.30}_{\pm0.49}$ & $71.46_{\pm0.36}$ & $78.10_{\pm0.51}$ & $89.32_{\pm1.63}$ & $90.07_{\pm0.60}$ & $92.05_{\pm0.78}$ & $\textbf{70.88}_{\pm0.35}$ & - \\
 & GLNN & $71.24_{\pm0.55}$ & $70.76_{\pm0.30}$ & $80.16_{\pm0.73}$ & $89.92_{\pm1.34}$ & $92.08_{\pm0.98}$ & $92.89_{\pm0.88}$ & $60.92_{\pm0.31}$ & - \\
 & KRD (ours) & $73.78_{\pm0.55}$ & $\textbf{71.80}_{\pm0.41}$ & $\textbf{81.48}_{\pm0.29}$ & $\textbf{90.37}_{\pm1.79}$ & $\textbf{93.15}_{\pm0.43}$ & $\textbf{93.86}_{\pm0.55}$ & $62.85_{\pm0.32}$ & - \\
 & \textit{Improv.} & 2.54 & 1.04 & 1.32 & 0.45 & 1.07 & 0.97 & 2.93 & 1.47 \\ \midrule
\multirow{4}{*}{GraphSAGE} & - & $\textbf{79.56}_{\pm0.47}$ & $70.24_{\pm0.62}$ & $79.40_{\pm0.48}$ & $89.76_{\pm1.51}$ & $89.96_{\pm0.56}$ & $91.79_{\pm0.69}$ & $\textbf{71.13}_{\pm0.32}$ & - \\
 & GLNN & $71.82_{\pm0.35}$ & $70.26_{\pm0.71}$ & $80.46_{\pm0.34}$ & $89.94_{\pm1.70}$ & $92.06_{\pm0.69}$ & $92.97_{\pm0.94}$ & $60.46_{\pm0.26}$ & - \\
 & KRD (ours) & $73.48_{\pm0.43}$ & $\textbf{70.94}_{\pm0.49}$ & $\textbf{81.36}_{\pm0.51}$ & $\textbf{90.37}_{\pm1.79}$ & $\textbf{92.96}_{\pm0.44}$ & $\textbf{93.91}_{\pm0.63}$ & $62.56_{\pm0.33}$ & - \\
 & \textit{Improv.} & 1.66 & 0.68 & 0.90 & 0.43 & 0.90 & 0.94 & 2.10 & 1.09 \\ \midrule
\multirow{4}{*}{GAT} & - & $\textbf{79.96}_{\pm0.63}$ & $69.58_{\pm0.43}$ & $79.02_{\pm0.43}$ & $90.54_{\pm1.73}$ & $90.50_{\pm0.97}$ & $91.99_{\pm1.08}$ & $\textbf{70.65}_{\pm0.23}$ & - \\
 & GLNN & $71.10_{\pm0.86}$ & $70.20_{\pm0.69}$ & $81.28_{\pm0.58}$ & $90.57_{\pm1.59}$ & $92.15_{\pm0.75}$ & $93.17_{\pm0.92}$ & $60.38_{\pm0.30}$ & - \\
 & KRD (ours) & $72.48_{\pm0.53}$ & $\textbf{70.64}_{\pm0.38}$ & $\textbf{82.00}_{\pm0.65}$ & $\textbf{91.10}_{\pm1.69}$ & $\textbf{93.23}_{\pm0.48}$ & $\textbf{94.02}_{\pm0.73}$ & $62.16_{\pm0.24}$ & - \\
 & \textit{Improv.} & 1.38 & 0.44 & 0.72 & 0.53 & 1.08 & 0.85 & 1.78 & 0.97 \\ \bottomrule
 
\end{tabular}} \vspace{-1.5em}
\end{center}
\end{table*}

\vspace{-0.5em}
\subsection{Classification Performance Comparison (Q1)}
\vspace{-0.1em}
The reliable knowledge of three teahcer GNNs is distilled into student MLPs in the transductive and inductive settings. The experimental results on seven datasets are reported in Table.~\ref{tab:1}, from which we can make three observations: \emph{(1)} Compared to the vanilla MLPs and intuitive KD baseline - GLNN, \texttt{KRD} performs significantly better than them in all cases, regardless of the datasets, teacher GNNs and evaluation settings. For example, \texttt{KRD} outperforms GLNN by 2.03\% (GCN), 1.92\% (SAGE), and 2.03\% (GAT) averaged over seven datasets in the transductive setting, respectively. The superior performance of \texttt{KRD} demonstrates the effectiveness of providing more reliable self-supervision for GNN-to-MLP distillation. \emph{(2)} The performance gain of \texttt{KRD} over GLNN is higher on the large-scale ogbn-arxiv dataset. We speculate that this is because the reliability of different knowledge points probably differ more in large-scale datasets, making those reliable knowledge points play a more important role. \emph{(3)} It can be seen that \texttt{KRD} works much better in the transductive setting than in the inductive one, since there are more node features that can be used for training in the transductive setting, providing more reliable knowledge points to serve as additional self-supervision.

\begin{table*}[!htbp]
\begin{center}
\caption{Performance comparison with leading graph distillation algorithms, where \textbf{bold} and \underline{underline} denote the best and second metrics. The experiments are conducted by adopting GCN as the teacher in the transductive setting (same for Table.~\ref{tab:3}, Table.~\ref{tab:4}, Fig.~\ref{fig:7}, and Fig.~\ref{fig:8}).}
\vspace{-0.5em}
\label{tab:2}
\resizebox{\textwidth}{!}{
\begin{tabular}{cc|ccccccc|c}

\toprule
\textbf{Category} & \textbf{Method} & \texttt{Cora} & \texttt{Citeseer} & \texttt{Pubmed} & \texttt{Photo} & \texttt{CS} & \texttt{Physics} & \texttt{ogbn-arxiv} & \textit{Avg. Rank} \\ \midrule
\multirow{2}{*}{Vanilla} & MLPs & $59.58_{\pm0.97}$ & $60.32_{\pm0.61}$ & $73.40_{\pm0.68}$ & $78.65_{\pm1.68}$ & $87.82_{\pm0.64}$ & $88.81_{\pm1.08}$ & $54.63_{\pm0.84}$ & 12.0 \\
 & Vanilla GCNs & $81.70_{\pm0.96}$ & $71.64_{\pm0.34}$ & $79.48_{\pm0.21}$ & $90.63_{\pm1.53}$ & $90.00_{\pm0.58}$ & $92.45_{\pm0.53}$ & $71.20_{\pm0.17}$ & 10.1 \\ \midrule
\multirow{5}{*}{GNN-to-GNN} & LSP & $82.70_{\pm0.43}$ & $72.68_{\pm0.62}$ & $80.86_{\pm0.50}$ & $91.74_{\pm1.42}$ & $92.56_{\pm0.45}$ & $92.85_{\pm0.46}$ & $71.57_{\pm0.25}$ & 7.4 \\
 & GNN-SD & $82.54_{\pm0.36}$ & $72.34_{\pm0.55}$ & $80.52_{\pm0.37}$ & $91.83_{\pm1.58}$ & $91.92_{\pm0.51}$ & $93.22_{\pm0.66}$ & $70.90_{\pm0.23}$ & 8.3 \\
 & TinyGNN & $83.10_{\pm0.53}$ & $73.24_{\pm0.72}$ & $81.20_{\pm0.44}$ & $92.03_{\pm1.49}$ & $93.78_{\pm0.38}$ & $93.70_{\pm0.56}$ & $72.18_{\pm0.27}$ & 4.7 \\
 & RDD & $83.68_{\pm0.40}$ & $73.64_{\pm0.50}$ & $\underline{81.74}_{\pm0.44}$ & $92.18_{\pm1.45}$ & $\textbf{94.20}_{\pm0.48}$ & $\underline{94.14}_{\pm0.39}$ & $\underline{72.34}_{\pm0.17}$ & 2.1 \\
 & FreeKD & $83.84_{\pm0.47}$ & $\underline{73.92}_{\pm0.47}$ & $81.48_{\pm0.38}$ & $\textbf{92.38}_{\pm1.54}$ & $93.65_{\pm0.43}$ & $93.87_{\pm0.48}$ & $\textbf{72.50}_{\pm0.29}$ & 2.9 \\ \midrule
\multirow{4}{*}{GNN-to-MLP} & GLNN & $82.20_{\pm0.73}$ & $71.72_{\pm0.30}$ & $80.16_{\pm0.20}$ & $91.42_{\pm1.61}$ & $92.22_{\pm0.72}$ & $93.11_{\pm0.39}$ & $67.76_{\pm0.23}$ & 9.7 \\
 & CPF & $83.56_{\pm0.48}$ & $72.98_{\pm0.60}$ & $81.54_{\pm0.47}$ & $91.70_{\pm1.50}$ & $93.42_{\pm0.48}$ & $93.47_{\pm0.41}$ & $69.05_{\pm0.18}$ & 6.4 \\
 & RKD-MLP & $82.68_{\pm0.45}$ & $73.42_{\pm0.45}$ & $81.32_{\pm0.32}$ & $91.28_{\pm1.48}$ & $93.16_{\pm0.64}$ & $93.26_{\pm0.37}$ & $69.87_{\pm0.25}$ & 7.3 \\
 & FF-G2M & $\underline{84.06}_{\pm0.43}$ & $73.85_{\pm0.51}$ & $81.62_{\pm0.37}$ & $91.84_{\pm1.42}$ & $93.35_{\pm0.55}$ & $93.59_{\pm0.43}$ & $69.64_{\pm0.26}$ & 4.9 \\
 & KRD (ours) & $\textbf{84.42}_{\pm0.57}$ & $\textbf{74.86}_{\pm0.58}$ & $\textbf{81.98}_{\pm0.41}$ & $\underline{92.21}_{\pm1.44}$ & $\underline{94.08}_{\pm0.34}$ & $\textbf{94.30}_{\pm0.46}$ & $70.92_{\pm0.21}$ & 2.1 \\ \bottomrule

\end{tabular}} \vspace{-1em}
\end{center}
\end{table*}

\vspace{-0.5em}
\subsection{Comparision with Representative Baselines (Q2)}
\vspace{-0.1em}
To answer \textbf{Q2}, we compare \texttt{KRD} with several representative graph knowledge distillation baselines, including both GNN-to-GNN and GNN-to-MLP distillation. As can be seen from the results reported in Table ~\ref{tab:2}, \texttt{KRD} outperforms all other GNN-to-MLP baselines by a wide margin. More importantly, we are the first work to demonstrate \emph{the promising potential of distilled MLPs to surpass distilled GNNs.} Even when compared with those state-of-the-art GNN-to-GNN distillation methods, \texttt{KRD} still shows competitive performance, ranking in the top two on 6 out of 7 datasets.

\subsection{Evaluation on Distribution Fitting Function (Q3)}
To evaluate the effectiveness of different distribution fitting functions and the momentum updating defined in Eq.~(\ref{equ:7}), we compare the learnable power distribution defined in Eq.~(\ref{equ:6}) with the other four schemes: (A) exponential distribution $p(s_i \mid \rho_i, \alpha) = \alpha\exp^{-\alpha\cdot\frac{\rho_i}{\rho_M}}$ with learnable rate $\alpha$; (B) Gaussian distribution $p(s_i \mid \rho_i, \alpha) = \mathcal{N}(0, \alpha)$ with learnable variance $\alpha$; (C) power distribution with fixed power $\alpha=1$; and (D) power distribution with fixed power $\alpha=3$. From the results reported in Table.~\ref{tab:3}, it can be seen that (1) when modeling the sampling probability with power distribution, the \emph{learnable} power is consistently better than the \emph{fixed} power on all datasets, and (2) the exponential, Gaussian and power distributions perform differently on different datasets, but the power distribution can achieve better overall performance than the other two distributions.

\begin{table}[!htbp]
\begin{center}
\vspace{-1em}
\caption{Performance comparison of different distribution fitting functions and the momentum updating of Eq.~(\ref{equ:7}), where \textbf{bold} and \underline{underline} denote the best and second metrics on each dataset.}
\vspace{-0.3em}
\label{tab:3}
\resizebox{\columnwidth}{!}{
\begin{tabular}{l|cccccc}

\toprule
\textbf{Methods} & \texttt{Cora} & \texttt{Citeseer} & \texttt{Pubmed} & \texttt{Photo} & \texttt{CS} & \texttt{Physics} \\ \midrule
Exponential & 83.30 & 73.84 & 81.10 & \underline{92.12} & 93.80 & 93.63 \\
Gaussian & \underline{84.12} & \underline{74.52} & \underline{81.56} & 92.10 & \textbf{94.15} & \underline{94.08} \\
Power (fixed $\alpha$=1) & 83.84 & 74.18 & 81.44 & 92.04 & 93.93 & 93.93 \\
Power (fixed $\alpha$=3) & 83.54 & 74.32 & 81.34 & 91.95 & 94.01 & 93.75 \\
Power (learnable) & \textbf{84.42} & \textbf{74.86} & \textbf{81.98} & \textbf{92.21} & \underline{94.08} & \textbf{94.30} \\ \bottomrule

\end{tabular}} \vspace{-1.2em}
\end{center}
\end{table}

\subsection{Evaluation on Knowledge Sampling Strategy (Q4)}
To explore how different sampling strategies influence the performance of distillation, we compare our knowledge-inspired sampling with other three schemes: (A) \textit{Non-sampling}: directly takes all nodes in the neighborhood as additional supervision and distills their knowledge into the student MLPs; (B) \textit{Random Sampling}: randomly sampling knowledge points with 50\% probability in the neighborhood for distillation; (C) \textit{Entropy-based Sampling}: performing min/max normalization on the information entropy of each knowledge point to [0-1], and then sampling by taking entropy as sampling probability. Besides, we also include the performance of vanilla GCN and GLNN as a comparison. We can observe from Table.~\ref{tab:3} that (1) Both non-sampling and random sampling help to significantly improve the performance of GLNN, again demonstrating the importance of providing additional supervision for training student MLPs. (2) Entropy- and knowledge-based sampling performs much better than non-sampling and random sampling, suggesting that different knowledge plays different roles during distillation. (3) Compared with entropy-based sampling, knowledge-based sampling fully takes into account the contextual information of the neighborhood as explained in Sec.~\ref{sec:4.2}, and thus shows better overall performance.

\begin{table}[!htbp]
\begin{center}
\vspace{-1em}
\caption{Performance comparison of different sampling strategies, where the best/second metrics are marked in \textbf{bold} and \underline{underline}.}
\vspace{-0.3em}
\label{tab:4}
\resizebox{\columnwidth}{!}{
\begin{tabular}{l|cccccc}

\toprule
\textbf{Methods} & \texttt{Cora} & \texttt{Citeseer} & \texttt{Pubmed} & \texttt{Photo} & \texttt{CS} & \texttt{Physics} \\ \midrule
Vanilla GCN & 81.70 & 71.64 & 79.48 & 90.63 & 90.00 & 92.45 \\
GLNN & 82.54 & 71.92 & 80.16 & 90.48 & 91.48 & 92.81 \\
Non-sampling & 83.26 & 73.58 & 80.74 & 91.45 & 93.04 & 93.42 \\
Random & 82.42 & 73.10 & 81.08 & 91.28 & 92.57 & \underline{93.74} \\
Entropy-based & \underline{83.64} & \underline{73.74} & \underline{81.32} & \underline{91.58} & \underline{93.35} & 93.63 \\ \midrule
Knowledge-based & \textbf{84.42} & \textbf{74.86} & \textbf{81.98} & \textbf{92.21} & \textbf{94.08} & \textbf{94.30} \\ \bottomrule

\end{tabular}} \vspace{-1.2em}
\end{center}
\end{table}

\subsection{Evaluation on Confidence Distribution (Q5)}
\begin{figure}[!htbp]
	\begin{center}
		\includegraphics[width=0.48\linewidth]{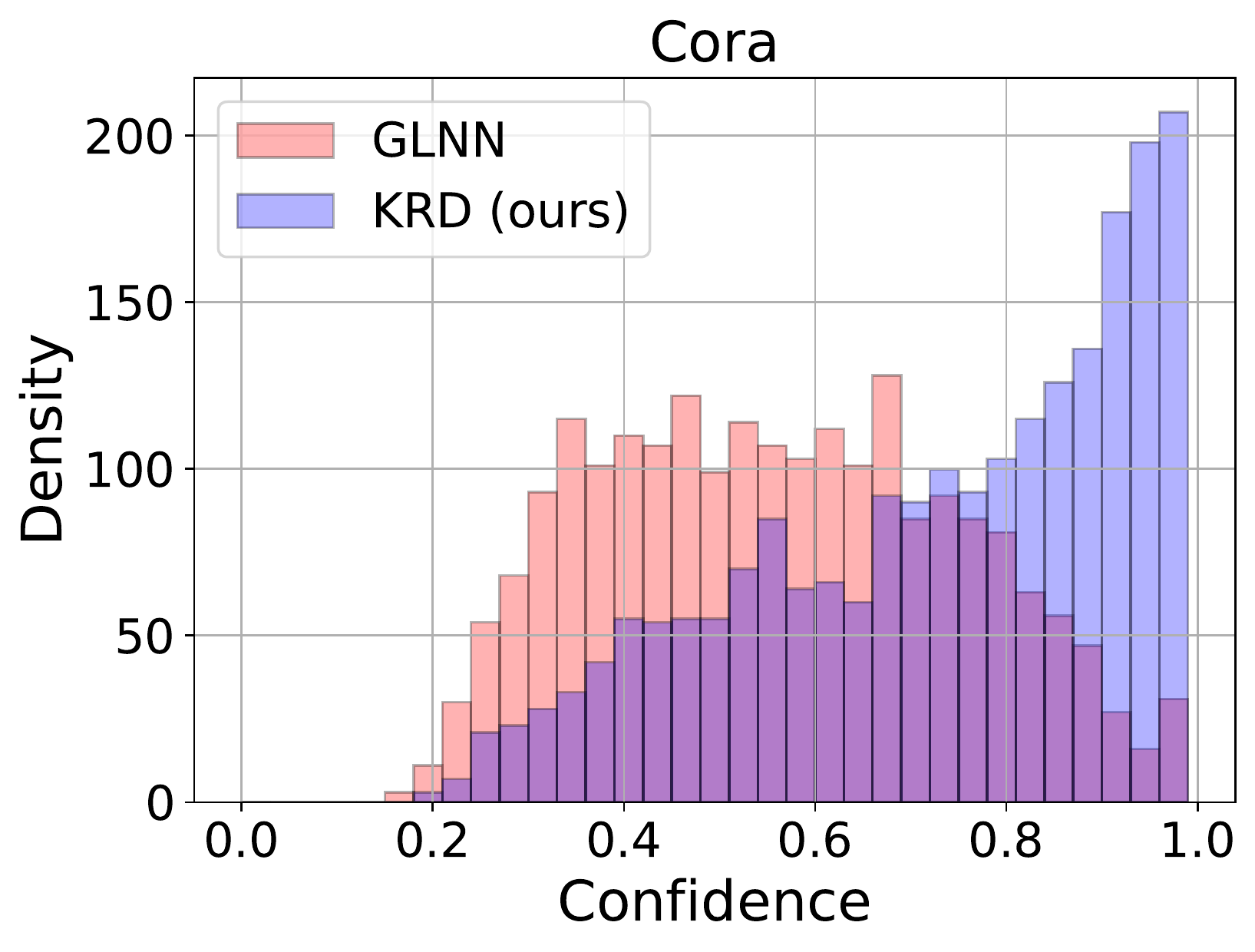}
		\includegraphics[width=0.48\linewidth]{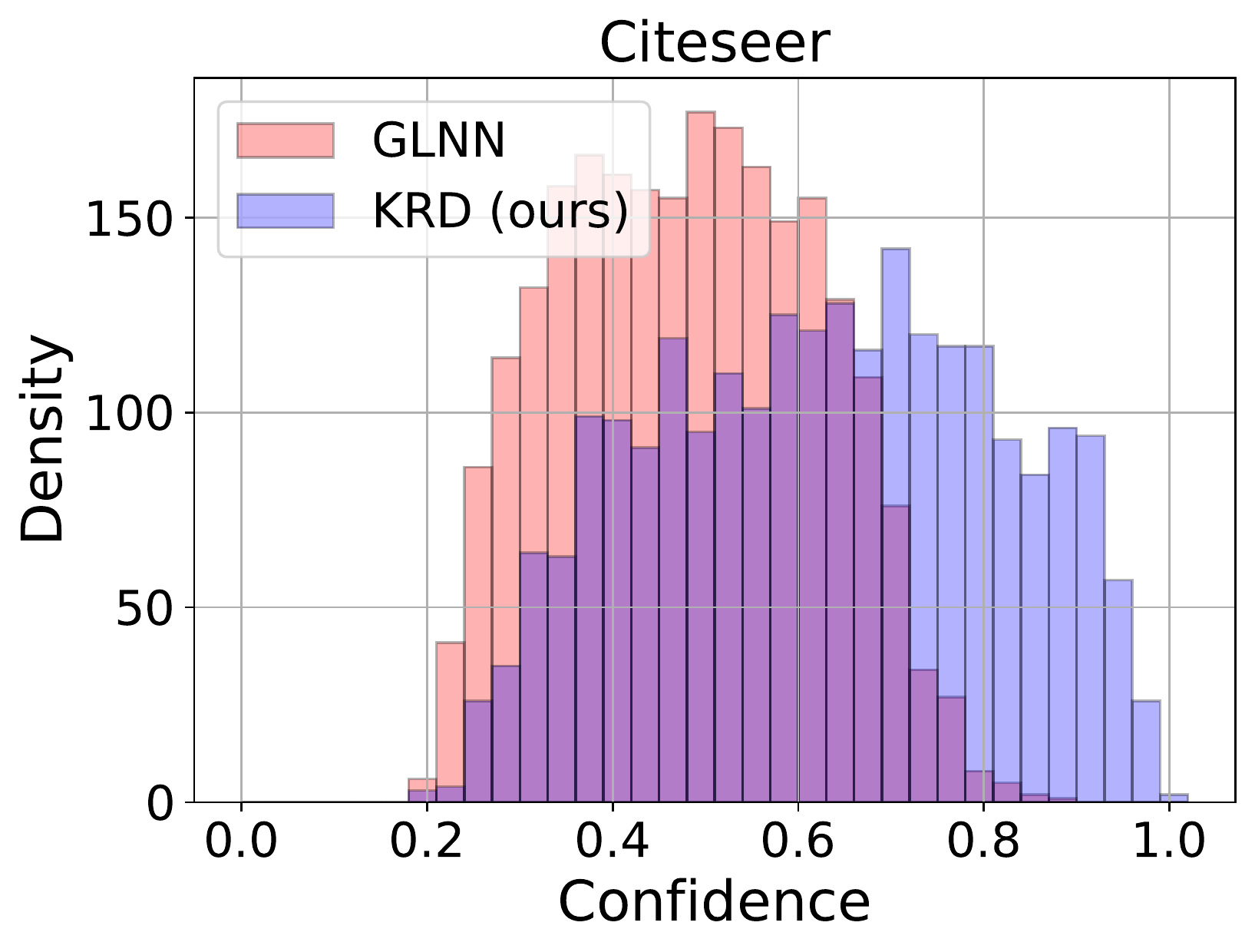}
		\includegraphics[width=0.48\linewidth]{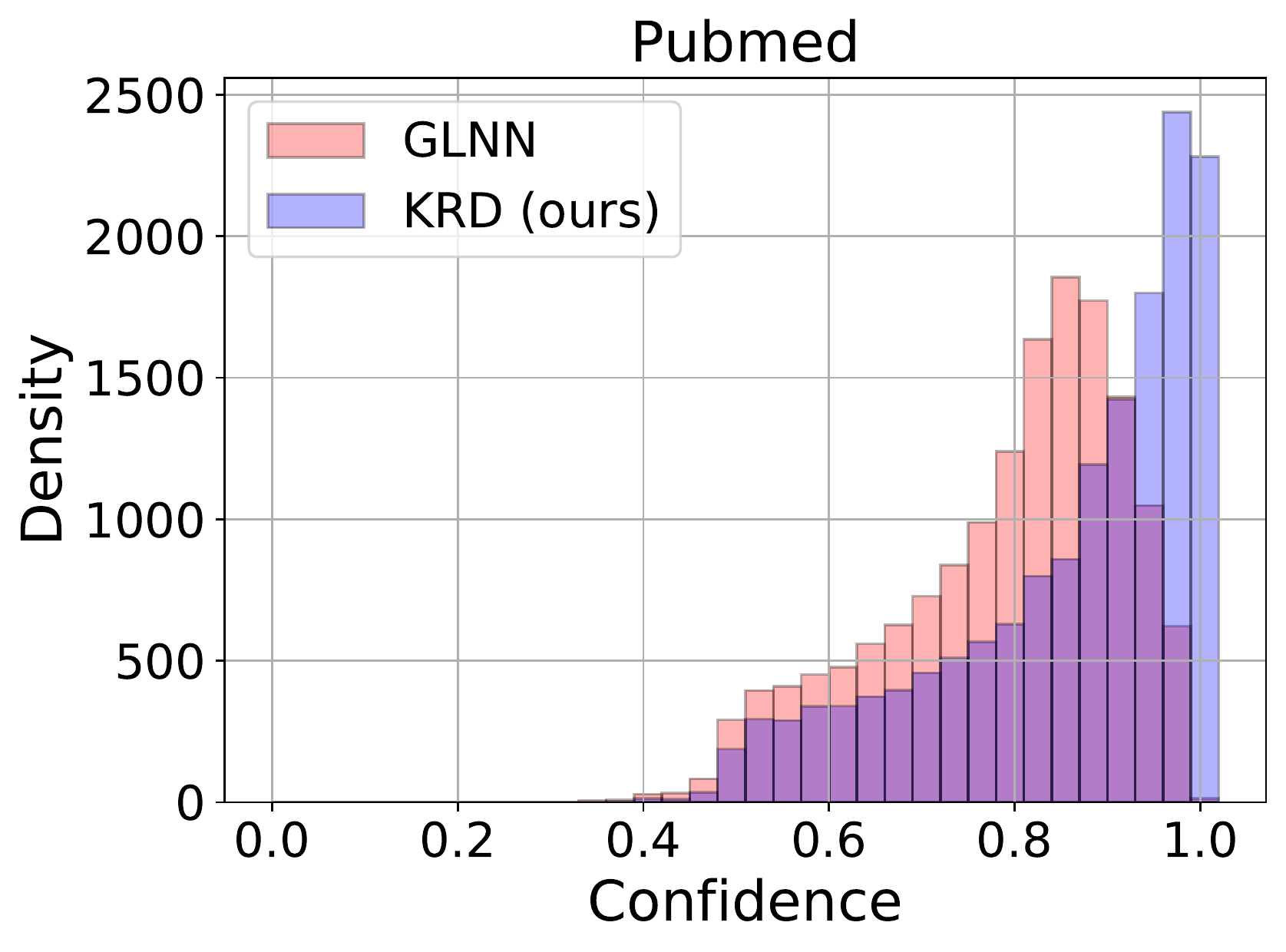} 
		\includegraphics[width=0.48\linewidth]{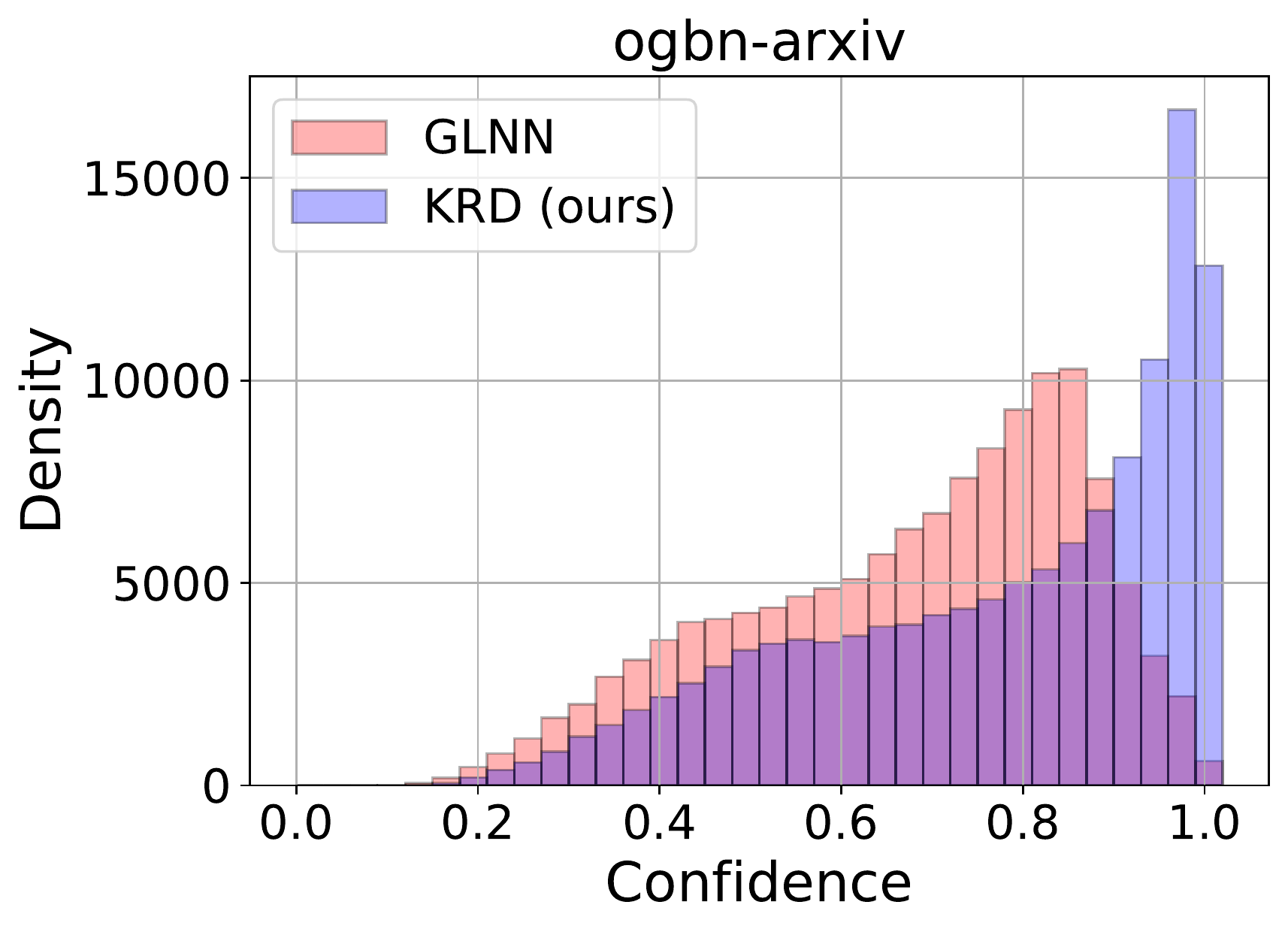}
	\end{center}
	\vspace{-1.2em}
	\caption{Confidence distribution of the distilled MLPs in GLNN and \texttt{KRD} on four datasets, where GCN is adopted as the teacher.}
	\vspace{-1em}
	\label{fig:7}
\end{figure}

To explore whether providing additional reliable supervision can improve the predictive confidence of distilled MLPs, we compare the confidence distribution of \texttt{KRD} with that of GLNN in Fig.~\ref{fig:7} on four datasets. It can be seen that the predictive confidence of student MLPs in GLNN (optimized with only the distillation term defined by Eq.~(\ref{equ:3}) is indeed not very high. Instead, \texttt{KRD} provides additional reliable self-supervision defined in Eq.~(\ref{equ:8}), which helps to greatly improve the predictive confidence of student MLPs.

\vspace{-0.8em}

\vspace{-0.5em}
\subsection{Evaluation on Hyperparameter Sensitivity (Q6)}
\begin{figure}[!htbp]
	\begin{center}
		\subfigure[Loss weight $\lambda$]{\includegraphics[width=0.48\linewidth]{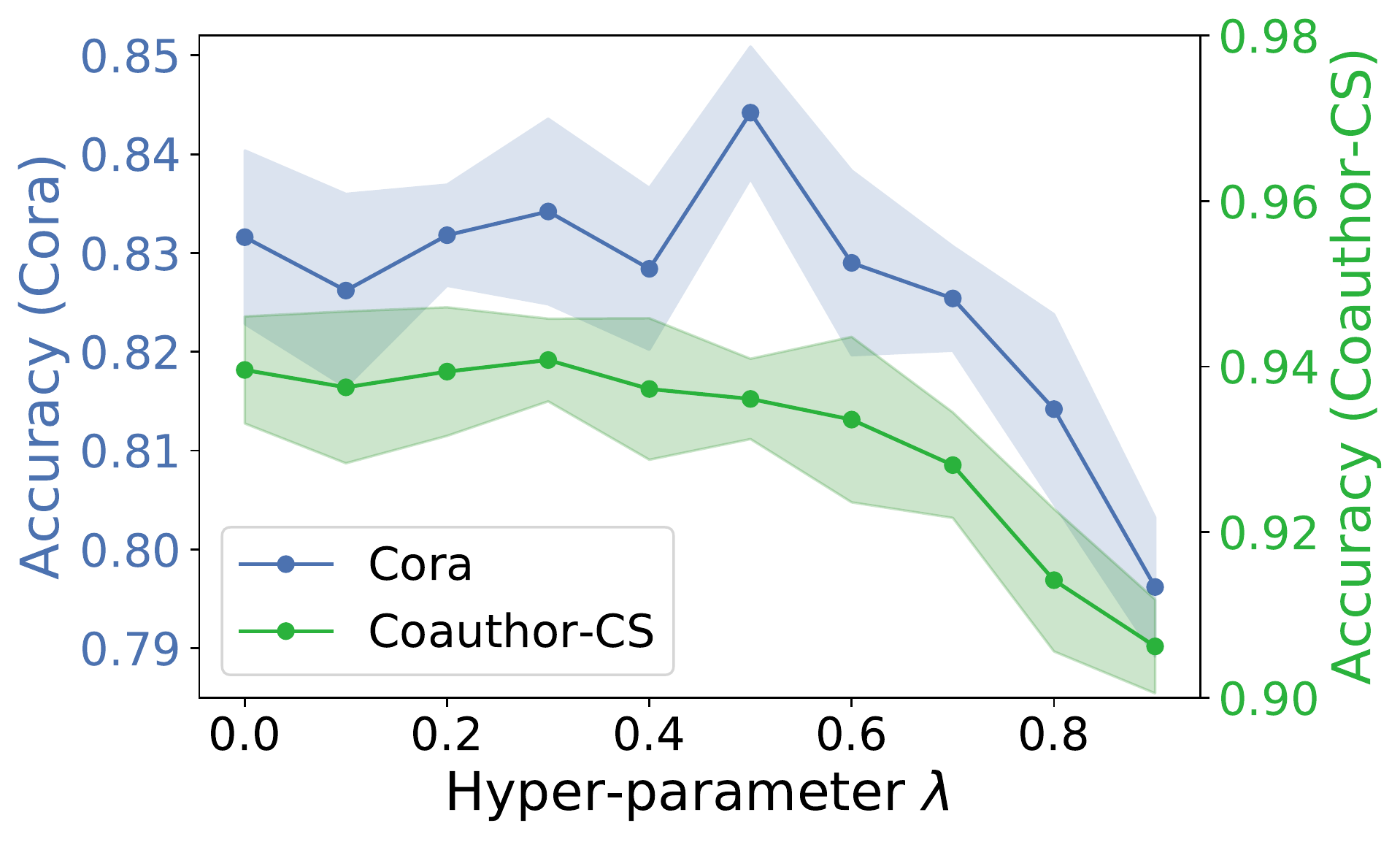} \label{fig:8a}}
		\subfigure[Momentum rate $\eta$]{\includegraphics[width=0.48\linewidth]{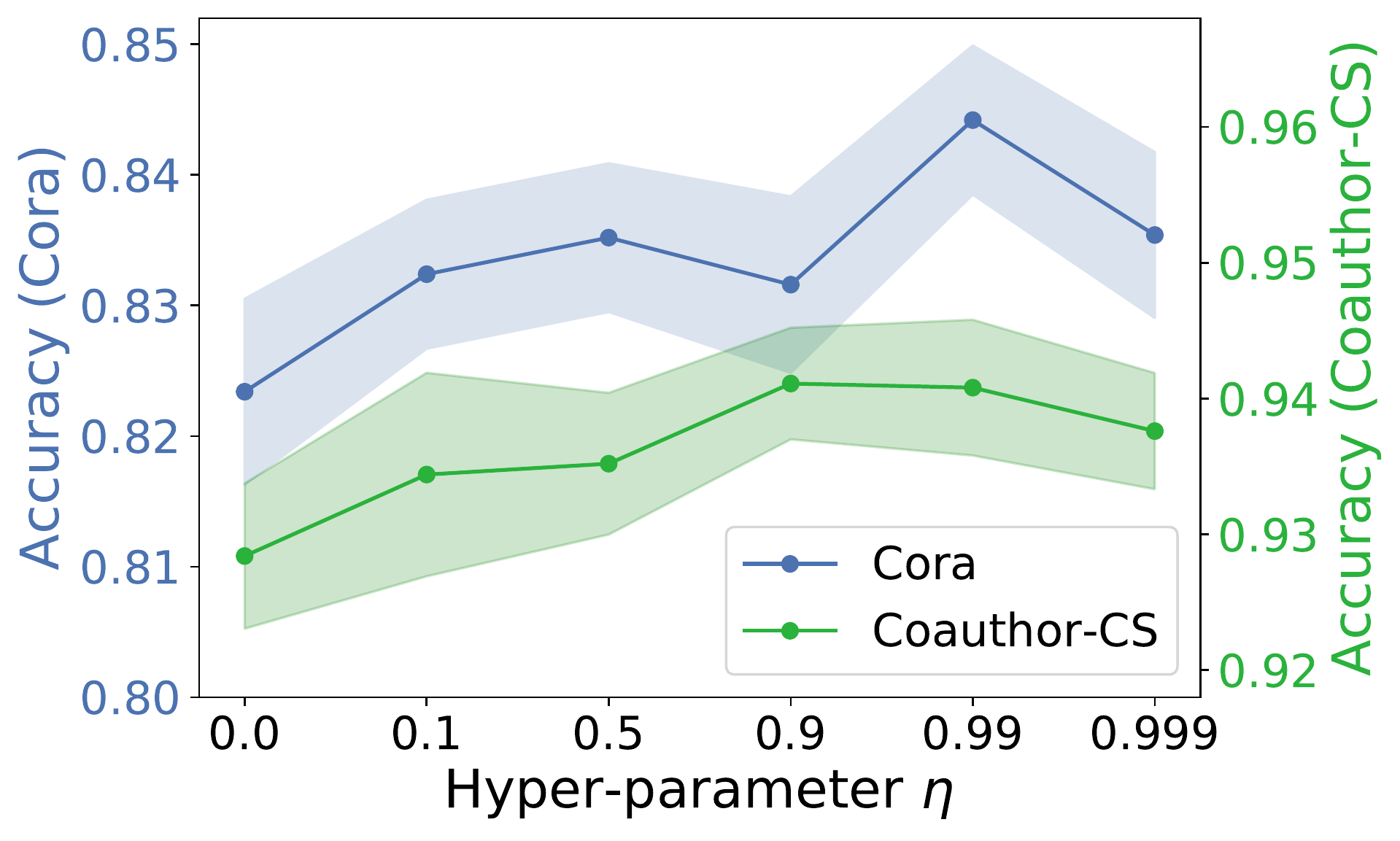} \label{fig:8b}}
	\end{center}
	\vspace{-1.2em}
	\caption{Hyperparameter sensitivity analysis on $\lambda$ and $\eta$.}
	\vspace{-0.5em}
	\label{fig:8}
\end{figure}

\vspace{-0.2em}
We provide sensitivity analysis for two hyperparameters, loss weights $\lambda$ and momentum updating rate $\eta$ in Fig.~\ref{fig:8a} and Fig.~\ref{fig:8b}, from which we observe that (1) setting the loss weight $\lambda$ too large weakens the contribution of the distillation term, leading to poor performance; (2) too large or small $\eta$ are both detrimental to modeling sampling probability and extracting informative knowledge. In practice, $\eta=0.9, 0.99$ often yields pretty good performance. In practice, we can determine $\lambda$ and $\eta$ by selecting the model with the highest accuracy on the validation set by the grid search.

\vspace{-0.8em}

\vspace{-0.8em}
\section{Conclusion}
\vspace{-0.2em}
In this paper, we identified a potential \emph{under-confidence} problem for GNN-to-MLP distillation, and more importantly, we described in detail what it represents, how it arises, what impact it has, and how to deal with it. To address this problem, we design a perturbation invariance-based metric to quantify the reliability of knowledge in GNNs, based on which we propose a \emph{Knowledge-inspired Reliable Distillation} (\texttt{KRD}) framework to make full use of those reliable knowledge points as additional supervision for training MLPs. Limitations still exist; for example, combining our work with other more powerful and expressive teacher/student models may be another promising direction.

\section{Acknowledgement}
This work was supported by National Key R\&D Program of China (No. 2022ZD0115100), National Natural Science Foundation of China Project (No. U21A20427), and Project (No. WU2022A009) from the Center of Synthetic Biology and Integrated Bioengineering of Westlake University.

\bibliography{example_paper}
\bibliographystyle{icml2023}

\clearpage
\renewcommand\thefigure{A\arabic{figure}}
\renewcommand\thetable{A\arabic{table}}
\renewcommand\theequation{A.\arabic{equation}}
\setcounter{table}{0}
\setcounter{figure}{0}
\setcounter{equation}{0}

\subsection*{A. Dataset Statistics} 

\emph{Seven} publicly available real-world graph datasets have been used to evaluate the proposed \texttt{KRD} framework. An overview summary of the statistical characteristics of these datasets is given in Table.~\ref{tab:A1}. For the three small-scale datasets, namely Cora, Citeseer, and Pubmed, we follow the data splitting strategy in \cite{kipf2016semi}. For the three large-scale datasets, including Coauthor-CS, Coauthor-Physics, and Amazon-Photo, we follow \cite{zhang2021graph,yang2021extract} to randomly split the data into train/val/test sets, and each random seed corresponds to a different data splitting. For the ogbn-arxiv dataset, we use the public data splits provided by the authors \cite{hu2020open}.

\begin{table}[!htbp]
\begin{center}
\vspace{-1.5em}
\caption{Statistical information of the datasets.}
\vspace{0.5em}
\label{tab:A1}
\resizebox{\columnwidth}{!}{
\begin{tabular}{lcccccccc}

\toprule
\textbf{Dataset} & \texttt{Cora} & \texttt{Citeseer} & \texttt{Pubmed} & \texttt{Photo} & \texttt{CS} & \texttt{Physics} & \texttt{ogbn-arxiv} \\ \midrule
\textbf{$\#$ Nodes} & 2708 & 3327 & 19717 & 7650 & 18333 & 34493 & 169343 \\
\textbf{$\#$ Edges} & 5278 & 4614 & 44324 & 119081 & 81894 & 247962 & 1166243 \\
\textbf{$\#$ Features} & 1433 & 3703 & 500 & 745 & 6805 & 8415 & 128 \\
\textbf{$\#$ Classes} & 7 & 6 & 3 & 8 & 15 & 5 & 40 \\
\textbf{Label Rate} & 5.2\% & 3.6\% & 0.3\% & 2.1\% & 1.6\% & 0.3\% & 53.7\% \\ \bottomrule

\end{tabular}} \vspace{-1.5em}
\end{center}
\end{table}

\begin{figure*}[!bp]
	\begin{center}
		\subfigure[Visualizations in GNNs]{\includegraphics[width=0.245\linewidth]{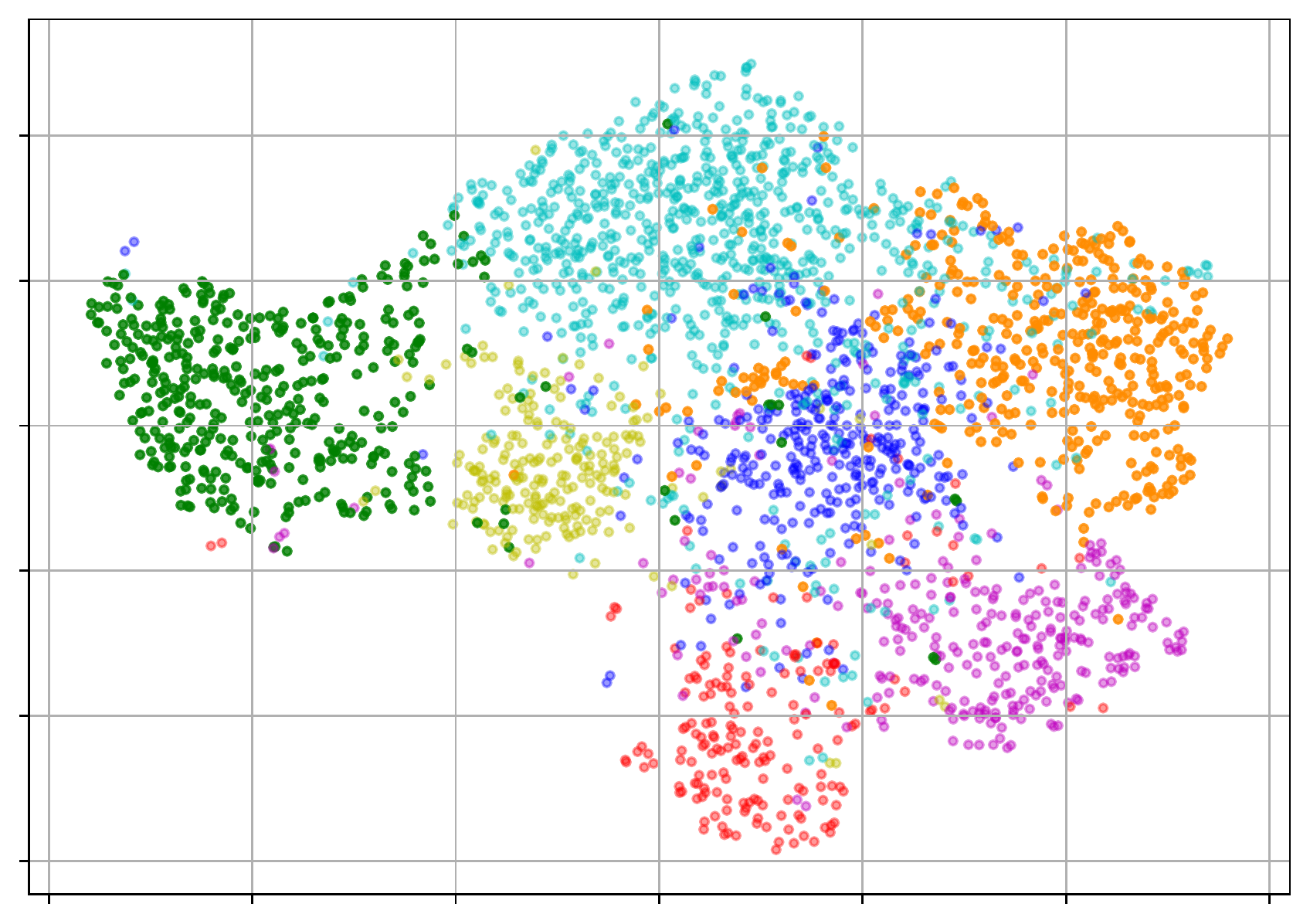}\label{fig:A1a}}
        \subfigure[Spatial Distribution in GNNs]{\includegraphics[width=0.245\linewidth]{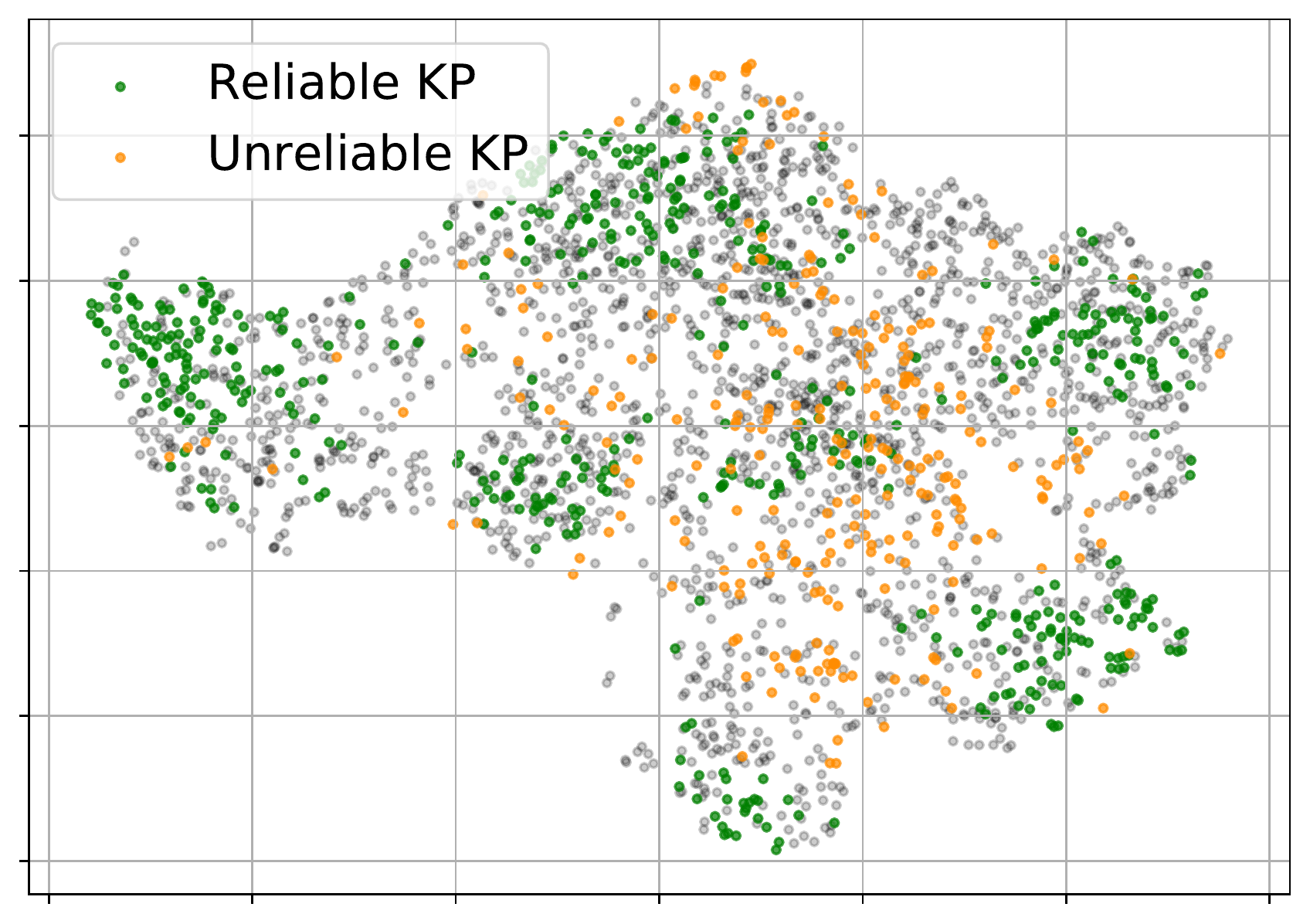}\label{fig:A1b}}
		\subfigure[Visualizations in MLPs]{\includegraphics[width=0.245\linewidth]{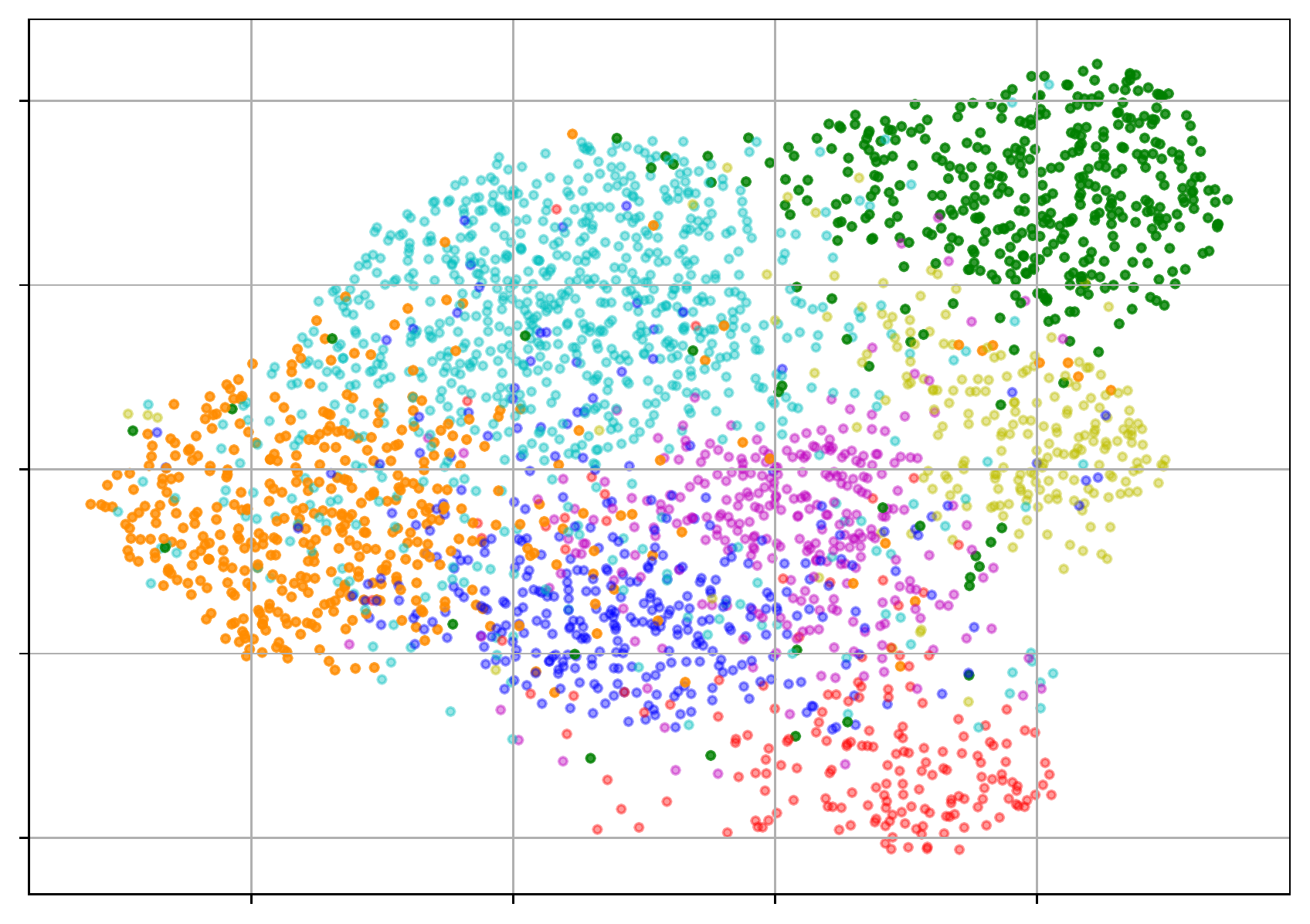}\label{fig:A1c}}
		\subfigure[Spatial Distribution in MLPs]{\includegraphics[width=0.245\linewidth]{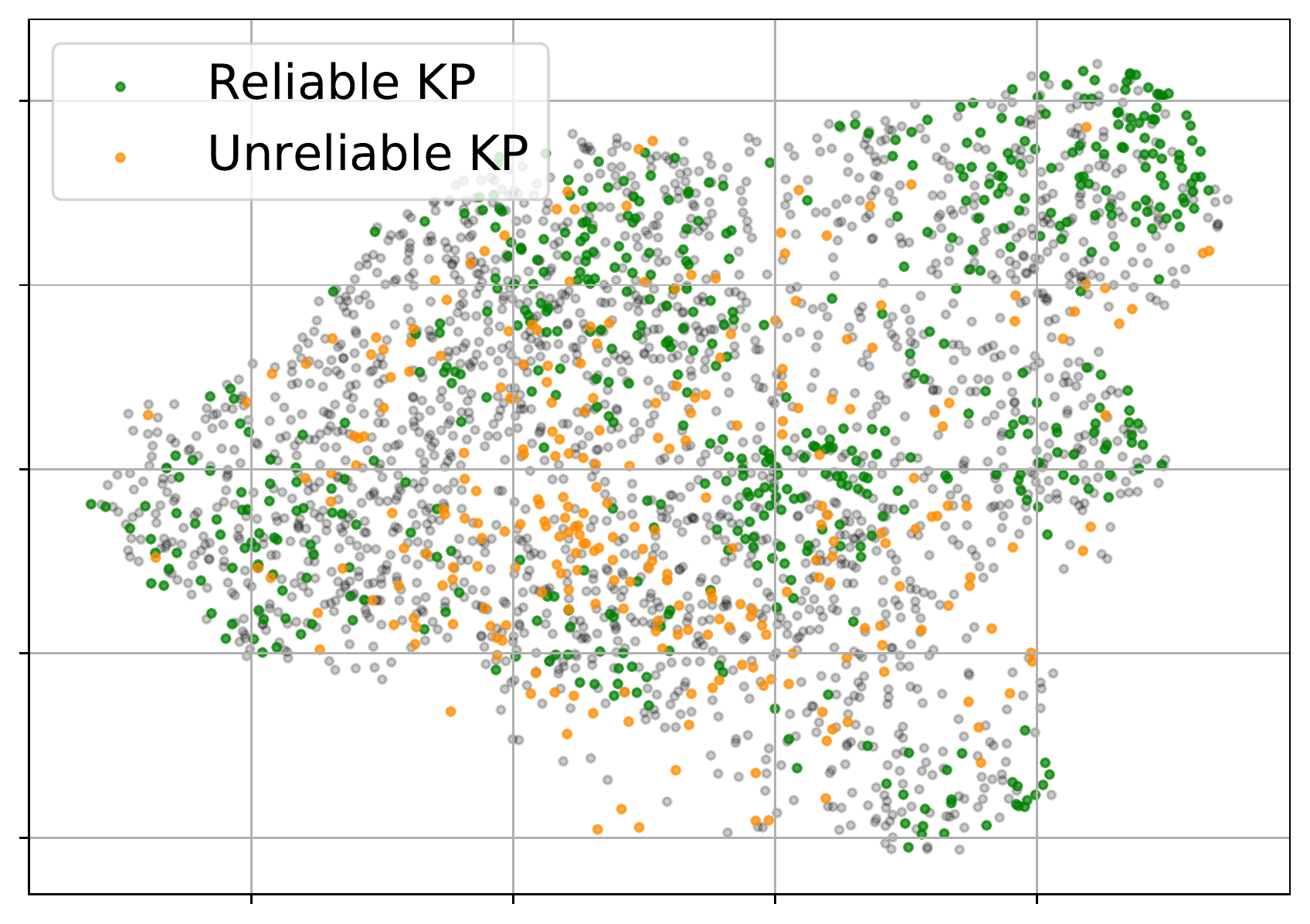}\label{fig:A1d}}
	\end{center}
	\vspace{-1em}
	\caption{\textbf{\emph{(a)(c)}} Visualizations of the embeddings of teacher GNNs and student MLPs for two classes on \texttt{Cora}. \textbf{\emph{(b)(d)}} Spatial distribution of knowledge points with the reliability ranked in the top 20\% and bottom 10\%, which are marked in {\color[rgb]{0.2156,0.7176,0.2773}green} and {\color[rgb]{1.0,0.6,0.0}orange}, respectively.}
	\vspace{-0.5em}
	\label{fig:A1}
\end{figure*}

\begin{figure*}[!bp]
	\begin{center}
		\subfigure[Cora]{\includegraphics[width=0.162\linewidth]{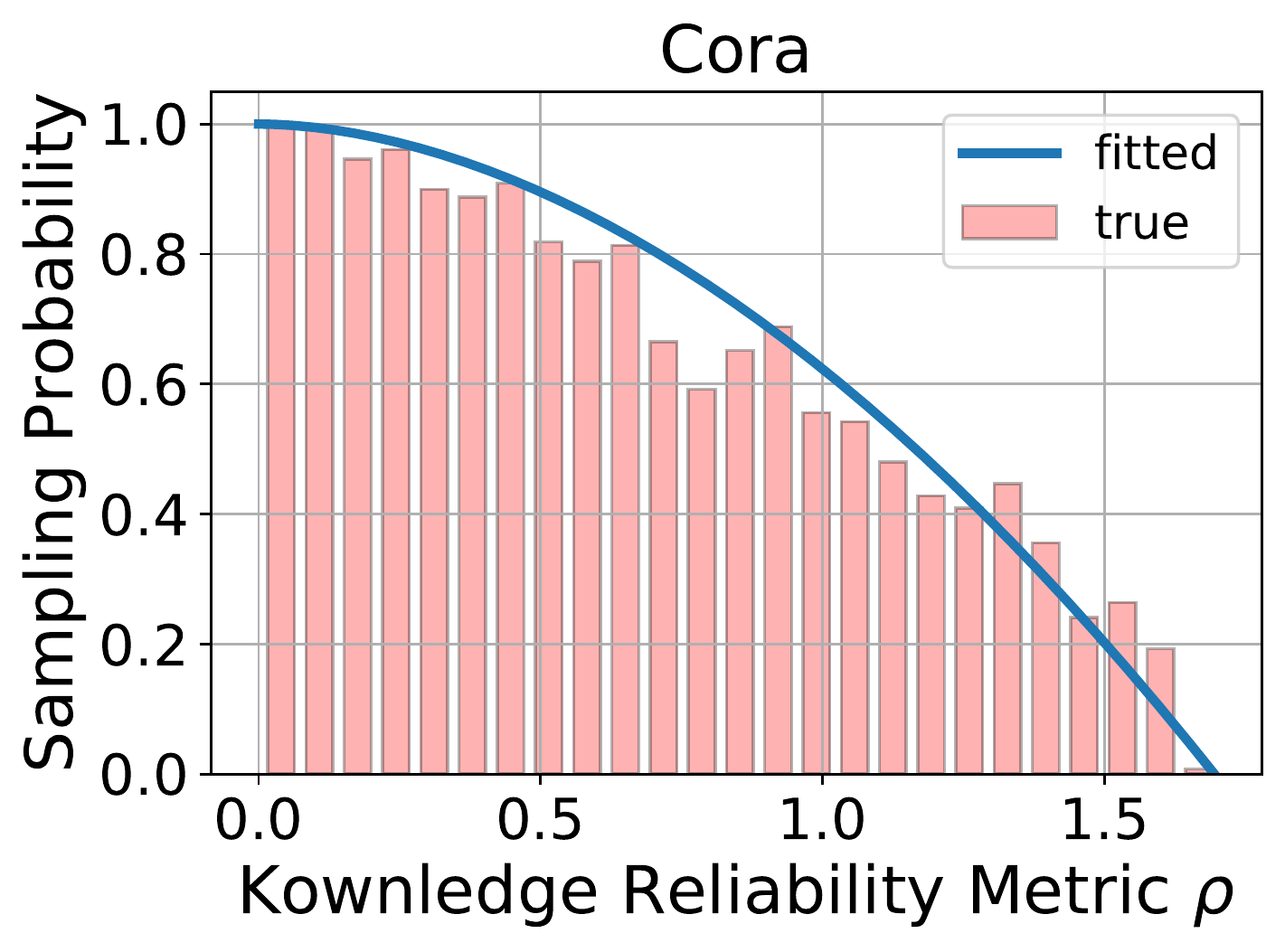}}
		\subfigure[Citeseer]{\includegraphics[width=0.162\linewidth]{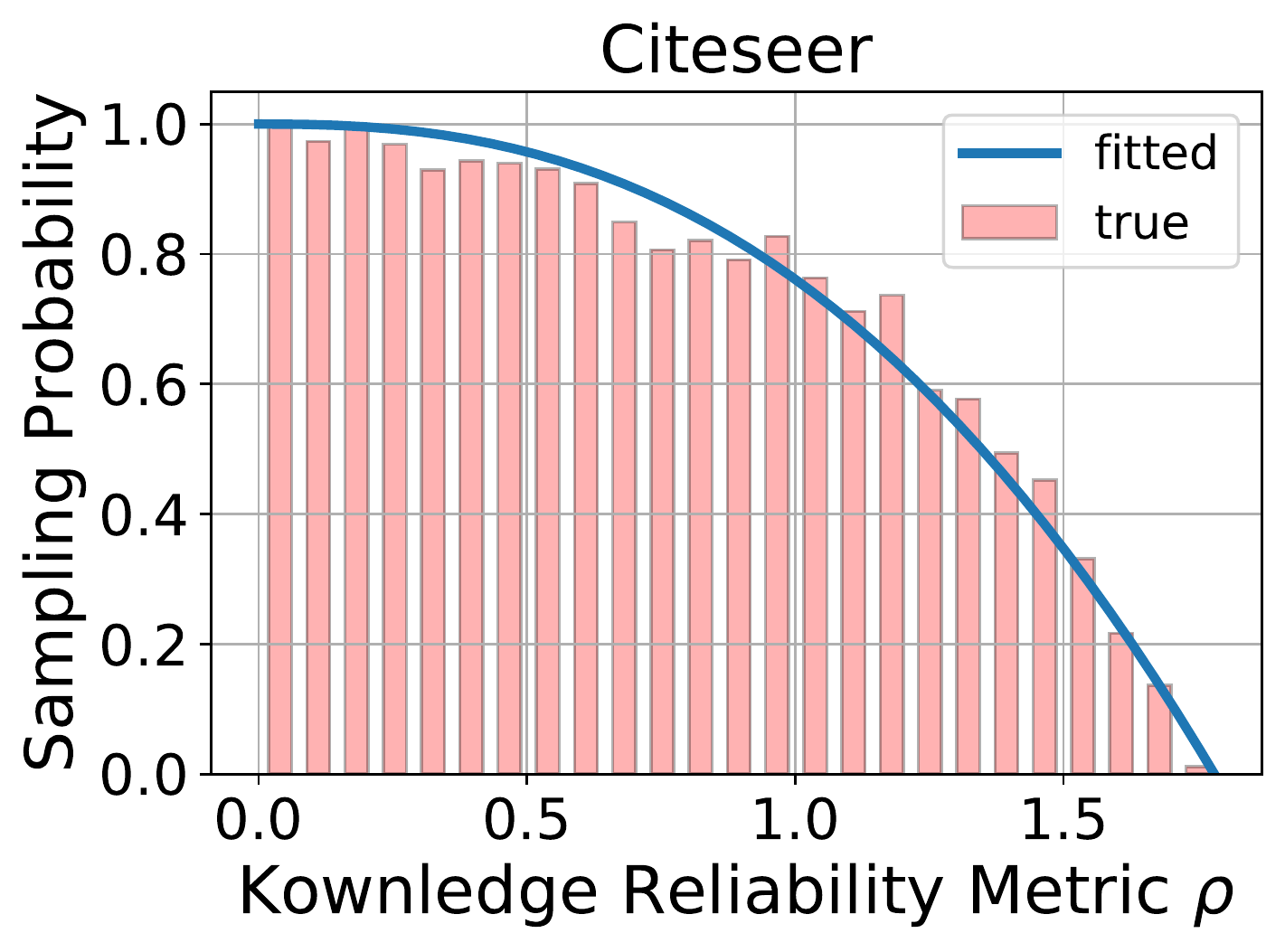}}
		\subfigure[Pubmed]{\includegraphics[width=0.162\linewidth]{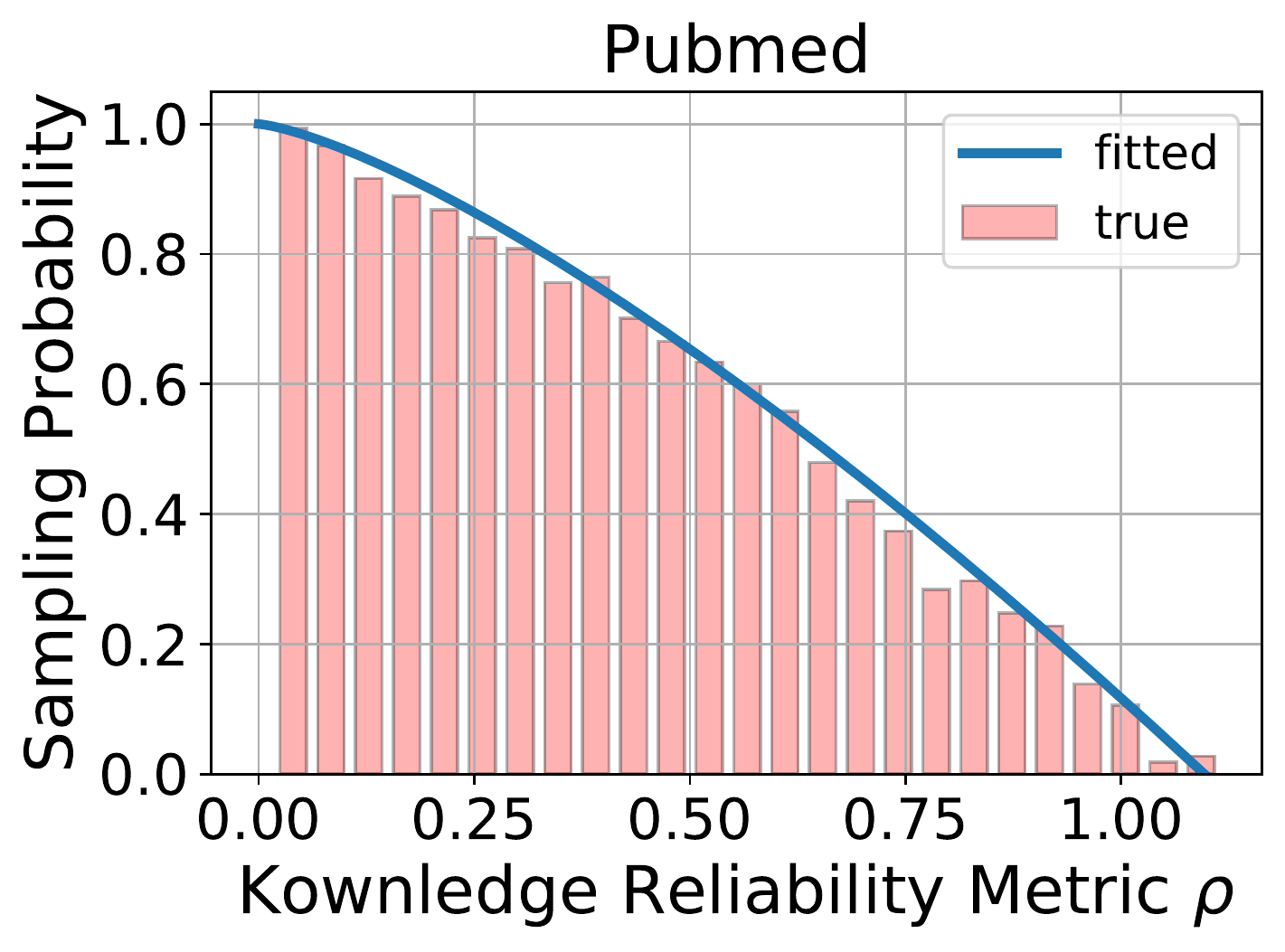}}
        \subfigure[Amazon-Photo]{\includegraphics[width=0.162\linewidth]{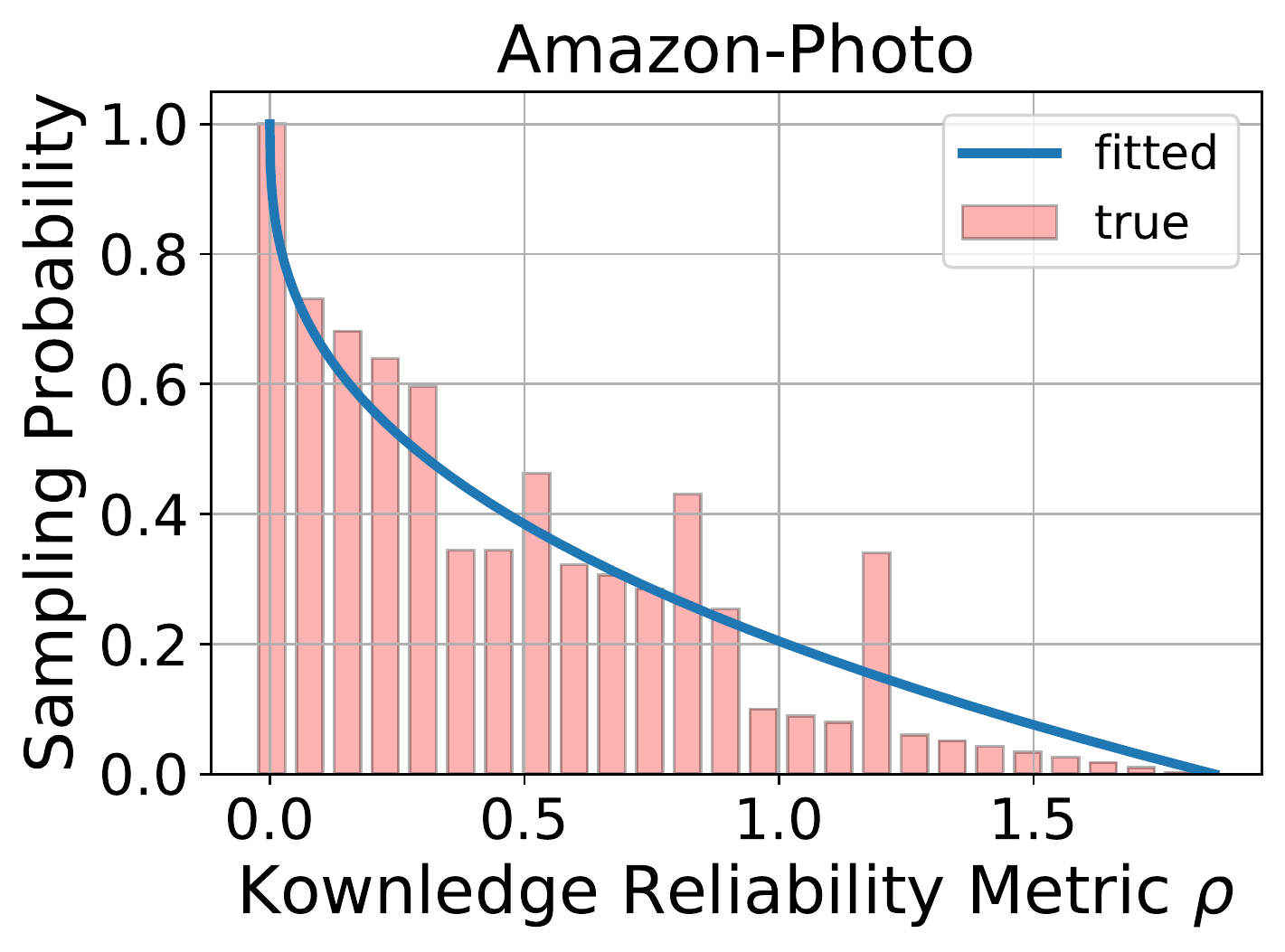}}
		\subfigure[Coauthor-CS]{\includegraphics[width=0.162\linewidth]{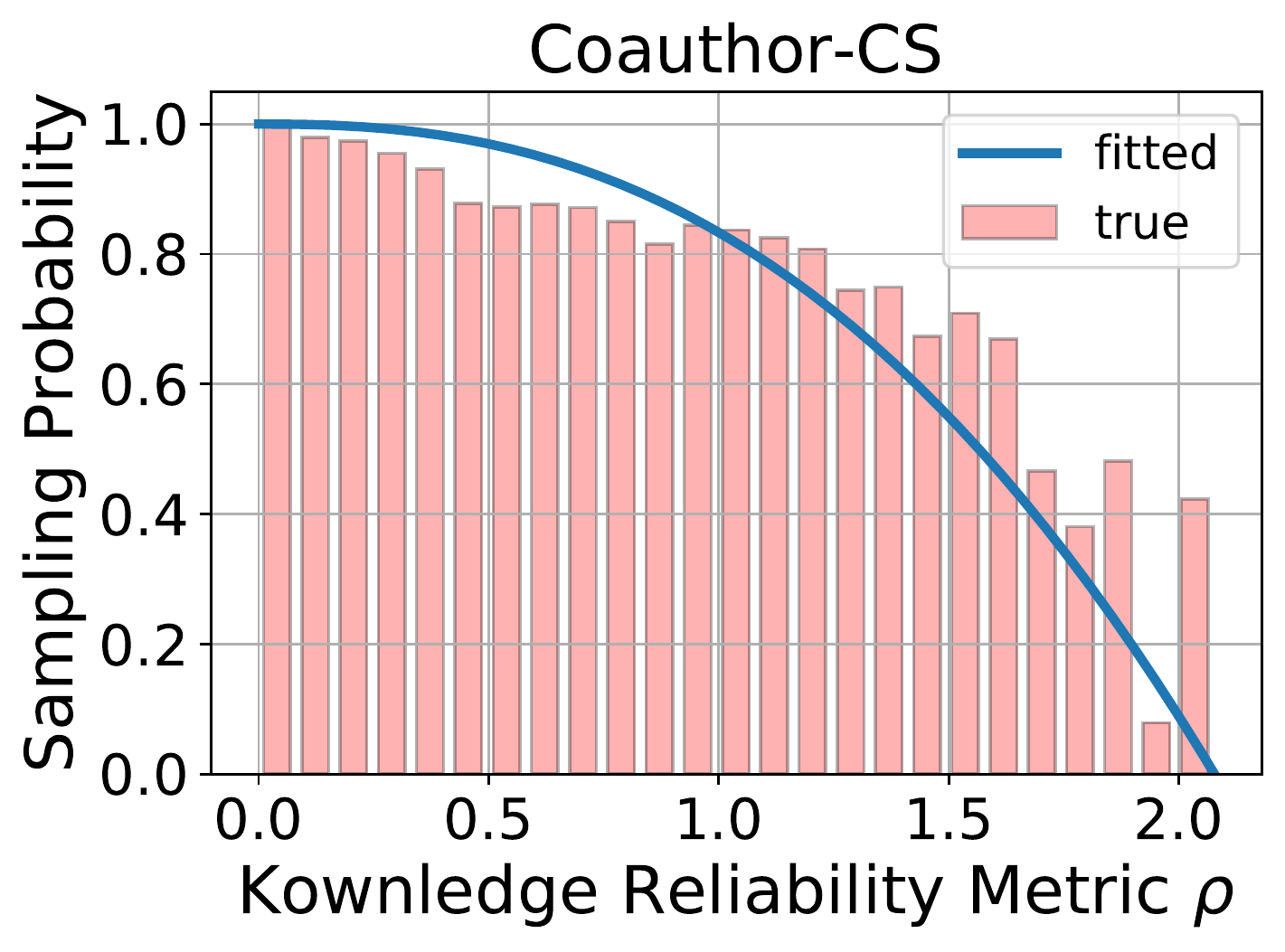}}
		\subfigure[Coauthor-Physics]{\includegraphics[width=0.162\linewidth]{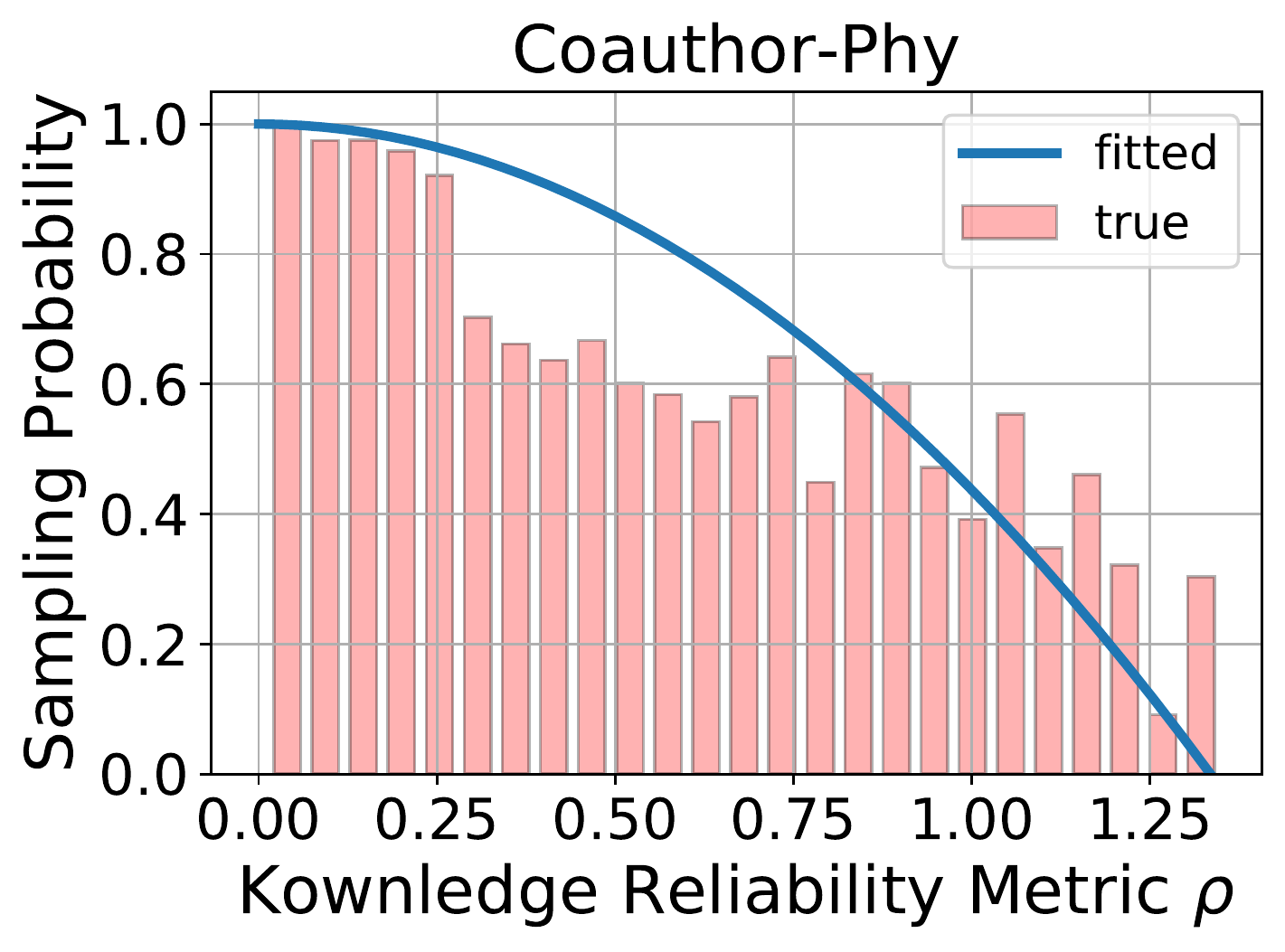}}
	\end{center}
	\vspace{-1em}
	\caption{Histograms of \emph{``True Positive"} sample density w.r.t the reliability metric $\rho$, as well as the fitted distributions (by learnable power distribution) of the sampling probability on six graph datasets, where the sample density has been min/max normalized to [0, 1].}
	\vspace{-0.5em}
	\label{fig:A2}
\end{figure*}

\subsection*{B. Implementation Details}
The following hyperparameters are set the same for all datasets: Epoch $E$ = 500, noise variance $\delta=1.0$, and momentum rate $\eta=0.99$ (0.9 for \texttt{ogb-arxiv}). The other dataset-specific hyperparameters are determined by an AutoML toolkit NNI with the hyperparameter search spaces as: hidden dimension $F=\{128, 256, 512, 1024, 2048\}$, layer number $L=\{2, 3\}$, distillation temperature $\tau=\{0.8, 0.9, 1.0, 1.1, 1.2\}$, loss weight $\alpha=\{0.0, 0.1, 0.2, 0.3, 0.4, 0.5\}$, learning rate $lr=\{0.001, 0.005, 0.01\}$, and weight decay $decay=\{0.0, 0.0005, 0.001\}$. For a fairer comparison, the model with the highest validation accuracy is selected for testing. Besides, the best hyperparameter choices of each setting are available in the supplementary. Moreover, the experiments on both baselines and our approach are implemented based on the standard implementation in the DGL library \cite{wang2019dgl} using the PyTorch 1.6.0 with Intel(R) Xeon(R) Gold 6240R @ 2.40GHz CPU and NVIDIA V100 GPU.

\textbf{Transductive \textit{vs.} Inductive.} We evaluate our model
under two evaluation settings: transductive and inductive. Their main difference is whether to use the test data for training. Specifically, we partition node features and labels into three disjoint sets, i.e., $\mathbf{X}=\mathbf{X}^L \sqcup \mathbf{X}_{obs}^U \sqcup \mathbf{X}_{ind}^U$,
and $\mathcal{Y}=\mathcal{Y}^L \sqcup \mathcal{Y}_{obs}^U \sqcup \mathcal{Y}_{ind}^U$. Concretely, the input and output of two settings are: (1) \textit{Transductive}: training on $\mathbf{X}$ and $\mathcal{Y}^L$ and testing on ($\mathbf{X}^U$, $\mathcal{Y}^U$). (2) \textit{Inductive}: training on $\mathbf{X}^L \sqcup \mathbf{X}_{obs}^U$ and $\mathcal{Y}^L$ and testing on ($\mathbf{X}_{ind}^U$, $\mathcal{Y}_{ind}^U$) \cite{anonymous2023double}.

\vspace{-0.5em}
\subsection*{C. More Results on Spatial Distribution}
The embeddings of teacher GNNs and student MLPs on the \texttt{Cora} dataset are visualized in Fig.~\ref{fig:A1}(a)(c). Then, we mark the knowledge points with the reliability ranked in the top 20\% and bottom 10\% as {\color[rgb]{0.2156,0.7176,0.2773}green} and {\color[rgb]{1.0,0.6,0.0}orange} in Fig.~\ref{fig:A1}(b)(d), respectively. It can be seen that most reliable knowledge points are distributed around the class centers, while those unreliable ones are distributed at the class boundaries.

\vspace{-0.5em}
\subsection*{D. More Results on Fitted Distributions}
We report histograms of \emph{``True Positive"} sample density w.r.t the reliability metric $\rho$ as well as the fitted distributions in Fig.~\ref{fig:A2}. It can be seen that the fitted distributions of the sampling probability closely matches the true histograms.

\end{document}